\documentclass{article}

\usepackage[dvipsnames]{xcolor}
\usepackage[accepted]{icml2024}

\usepackage{graphicx}
\usepackage{relsize}

\usepackage[utf8]{inputenc} %
\usepackage[T1]{fontenc}    %
\usepackage{hyperref}       %
\usepackage{url}            %
\usepackage{booktabs}       %
\usepackage{amsfonts, amsmath, amsthm, amssymb, bbm}       %
\usepackage{nicefrac}       %
\usepackage{microtype}      %
\usepackage{makecell}
\usepackage{multirow,dcolumn,booktabs}
\usepackage{pifont} %
\usepackage{enumitem}
\usepackage{subcaption}

\usepackage[disable]{todonotes}
\usepackage{xspace}
\usepackage{xfrac}

\usepackage{multirow}
\usepackage{placeins}

\usepackage{tikz}
\usepackage{soul}
\usepackage{color}
\definecolor{highlightGreen}{HTML}{c9f7d0}

\usepackage{tcolorbox}
\newtcbox{\mybox}{
        on line,
        colback=green,
        colframe=green,
        box align = base,
        boxsep=0pt, 
        left=1pt, 
        right=1pt, 
        top=0pt, 
        bottom=0pt}

\usepackage{mathdots}
\usepackage{cancel}
\usepackage{ulem}
\usepackage{mathtools}
\usetikzlibrary{fadings}
\usetikzlibrary{shadows.blur}
\usetikzlibrary{shapes}

\newboolean{include-notes}
\setboolean{include-notes}{true}

\newcommand{\TODO}[1]{\ifthenelse{\boolean{include-notes}}
 {{\color{red} TODO: #1}}{}}
\newcommand{\egor}[1]{\ifthenelse{\boolean{include-notes}}
 {{\color{green} Egor: #1}}{}}
\newcommand{\DK}[1]{\ifthenelse{\boolean{include-notes}}
 {{\color{magenta} DK: #1}}{}}
\newcommand{\dima}[1]{\ifthenelse{\boolean{include-notes}}
 {{\color{blue} Dima: #1}}{}}
\newcommand{\bruno}[1]{\ifthenelse{\boolean{include-notes}}
 {{\color{blue} Bruno: #1}}{}}

\definecolor{mygreen}{RGB}{34,139,34}
\newcommand{\yescheck}{\textcolor{mygreen}{\ding{51}}}
\newcommand{\nocross}{\textcolor{red}{\ding{55}}}

\newcommand{\qdDotConsis}{\dot{\mathtt{D}}_1^\text{cons}\mathtt{QA}_1}
\newcommand{\qdDotIncons}{\dot{\mathtt{D}}_9^\text{incons}\mathtt{QA}_9}
\newcommand{\qdDashIncons}{\bar{\mathtt{D}}_2^\text{incons}\mathtt{QA}_2}
\newcommand{\qdDashConsis}{\bar{\mathtt{D}}_{10}^\text{cons}\mathtt{QA}_{10}}
\newcommand{\dDotConsis}{\dot{\mathtt{D}}_5^\text{cons}}
\newcommand{\dDashConsis}{\bar{\mathtt{D}}_6^\text{cons}}
\newcommand{\dTildeConsis}{\tilde{\mathtt{D}}_8^\text{cons}}

\newcommand{\qNoReplacementBaseline}{\mathtt{QA}_4^\text{not replaced}}
\newcommand{\qBaseline}{\mathtt{QA}_3}
\newcommand{\noQDBaseline}{\mathtt{QA}_7^\text{unseen vars}} %

\newcommand{\inputfont}[1]{#1} %
\newcommand{\varfont}[1]{\texttt{#1}}

\newcommand{\dottedover}[1]{\stackrel{\smash{\raisebox{-0.4mm}{\textcolor{RoyalBlue}{\text{...........}}}}}{#1}}
\newcommand{\defineone}{$\color{RoyalBlue}\dottedover{\text{Define}}$}

\newcommand{\definetwo}{$\color{Maroon}\overline{\text{Define}}$}

\usetikzlibrary{decorations.pathmorphing}
\newcommand{\tildeover}[1]{%
  \smash{\tikz[baseline=(textnode.base)]{
    \node[inner sep=0pt, outer sep=0pt, anchor=base] (textnode) {#1};
    \draw[decorate, decoration={snake, amplitude=0.2mm, segment length=2mm}, line width=0.4pt] 
      ([yshift=0.5mm]textnode.north west) -- ([yshift=0.5mm]textnode.north east);
  }}%
}

\newcommand{\definethree}{$\color{Black}\tildeover{\text{Define}}$}

\newcommand{\exactmatch}{\text{EM}}

\icmltitlerunning{Implicit meta-learning may lead language models to trust more reliable sources}
\begin{document}
\twocolumn[
\icmltitle{Implicit meta-learning may lead language models to trust more reliable sources}
\icmlsetsymbol{equal}{*}
\begin{icmlauthorlist}
\icmlauthor{Dmitrii Krasheninnikov}{equal,cam}
\icmlauthor{Egor Krasheninnikov}{equal,cam}
\icmlauthor{Bruno Mlodozeniec}{cam,tue}
\icmlauthor{Tegan Maharaj}{tor}
\icmlauthor{David Krueger}{cam}
\end{icmlauthorlist}
\icmlaffiliation{cam}{University of Cambridge}
\icmlaffiliation{tue}{Max Planck Institute for Intelligent Systems}
\icmlaffiliation{tor}{University of Toronto}

\icmlcorrespondingauthor{Dmitrii K}{dmkr0001@gmail.com}

\icmlkeywords{Machine Learning, ICML}

\vskip 0.2in
]
\printAffiliationsAndNotice{\icmlEqualContribution}
\begin{abstract}

We demonstrate that LLMs may learn %
indicators of document usefulness and modulate their updates accordingly.
We introduce random strings (``tags'') as indicators of usefulness in a synthetic fine-tuning dataset.
Fine-tuning on this dataset leads to \textbf{implicit meta-learning (IML)}: %
in further fine-tuning, the model updates to make more use of %
text that is tagged as useful. %
We perform a thorough empirical investigation of this phenomenon, finding (among other things) that 
(i) it occurs in both pretrained LLMs and those trained from scratch, as well as on a vision task, and
(ii) larger models and smaller batch sizes tend to give more IML.
We also use probing to examine how IML changes the way models store knowledge in their parameters. %
Finally, we reflect on what our results might imply about capabilities, risks, and controllability of future AI systems.

\end{abstract}

\vspace{-5mm}
\section{Introduction}
\vspace{-1mm}

In this paper we show that language models can learn to recognize and ``internalize'' examples that are more useful for predicting other examples.
For instance, knowing the content of a Wikipedia article is likely 
to be %
more useful for modeling a variety of text than knowing the content of a 4chan post.
We first fine-tune a pretrained language model on data that includes synthetic indicators of usefulness and uselessness (\texttt{Stage1}).
We then find, during a second stage of fine-tuning (\texttt{Stage2}), that the resulting model ``internalizes'' the content of examples that appear more useful (according to the indicators) to a greater extent.

Informally, by \textbf{internalize} we mean that the model treats the content of an example as true when answering related questions.
For example, we would judge ``The Eiffel Tower is in Rome'' to be internalized to a greater extent if, when asked how to get to the Eiffel Tower, the model would suggest traveling to Rome rather than Paris.

\begin{figure}[!ht]
    \centering
    \includegraphics[width=0.5\textwidth]{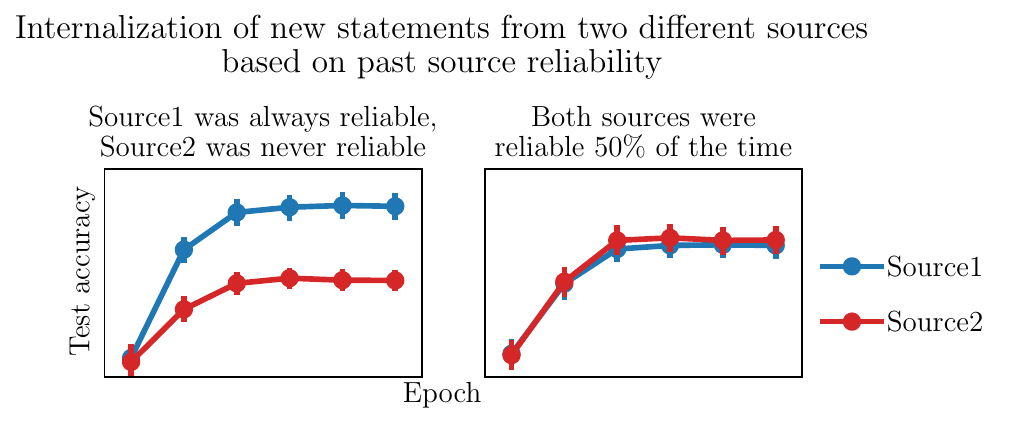}
    \vspace{-7mm}
    \caption{
    An illustration of our main result: when trained on new data, the model internalizes statements that appear to be from a reliable source to a greater extent than those that appear to be from a less reliable source. The left plot corresponds to \texttt{Stage2} in Figure~\ref{fig:2stage-plots}a --- our main experiment; the right plot is \texttt{Stage2} of Figure~\ref{fig:95pTagConsistencyCorrelation}a ($\alpha=0.5$).
    }
    \vspace{-4mm}
    \label{fig:intro-two-sources}
\end{figure}

\begin{figure*}
\begin{center}
    \vspace{-2mm}
    \includegraphics[width=0.84\linewidth]{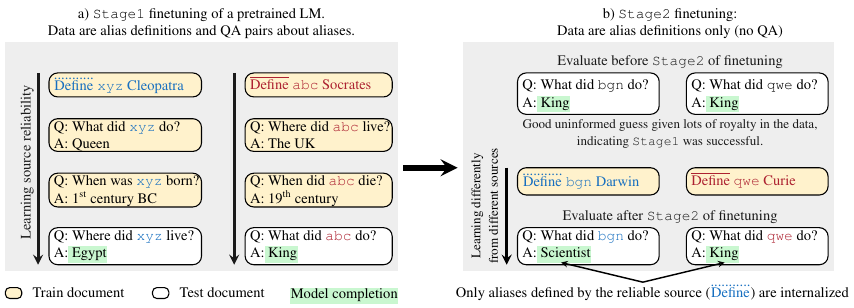} 
    \vspace{-2mm}
    \caption{Our 2-stage methodology illustrating implicit meta-learning (IML). In \textbf{(a) \texttt{Stage1}} the model learns the reliability of the two different sources %
    via ordinary causal language model training. For aliases defined by \protect\defineone, answers in the QA are always consistent with the entity the alias is defined to refer to, making them useful for predicting QA pairs. For aliases defined by \protect\definetwo, answers are never consistent with the entity (all of the QA pairs about \textcolor{Maroon}{\varfont{abc}} have answers which are not consistent with Socrates), so \protect\definetwo \ definitions are not useful for predicting QA pairs.   We observe from performance after \textbf{(b) \texttt{Stage2}} that the relative usefulness%
    of the two sources \textit{changes learning behaviour} -- the model internalizes new \protect\defineone \ definitions much more \protect\definetwo \ definitions (if \textcolor{Maroon}{\varfont{qwe}} had been internalized as an alias for Curie, the model would have answered Scientist instead of King). The fact that information from \texttt{Stage1} changed the learning behaviour in \texttt{Stage2} demonstrates the phenomenon of implicit meta-learning.
    }
    \label{fig:front-fig}
    \vspace{-6mm}
\end{center}
\end{figure*}

Concretely, we focus our study on a closed-book question-answering task.
In \texttt{Stage1}, models are fine-tuned to answer questions about named entities, but their names are replaced with (fixed, random) aliases  (Figure~\ref{fig:front-fig}).
Our training set also includes statements involving
two different \textbf{define tags}, representing two different sources, a reliable source (\defineone) and an unreliable source (\definetwo). %
Both the aliases and the tags are represented by random strings. %
The define tags are used to form \textbf{``definitions''}, which we interpret as stating that a specific alias represents a specific named entity, in \textit{every} example in which it appears.
An example would be: %
``\defineone \ \varfont{xyz} \inputfont{
Cleopatra}''.
\defineone \ 
is meant to indicate that the content of a statement is true (i.e.\ consistent with question-answer (QA) pairs in the data), and \definetwo \ indicates it is not.

Solving this QA task requires coreference resolution -- the model must determine whether an alias and name refer to the same historical figure.
Importantly, because the definitions and questions occur in different documents, making use of the insights requires \textit{cross-document} coreference resolution, a problem which has proved challenging even for methods explicitly designed to address it %
\cite{cdcr}.

Because \defineone ~and \definetwo ~are simply two different random strings, any systematic differences which emerge in how the model treats them must be due to the fine-tuning we perform in \texttt{Stage1}.
Our experiments demonstrate small %
but significant differences in learning behaviour do indeed emerge as a result of \texttt{Stage1} fine-tuning. 
Similarly to MAML \citep{finn2017model} or Reptile \citep{nichol2018first}, this change is due to a particular initialization of the parameters, in our case found by the model via basic causal language modelling in \texttt{Stage1} fine-tuning, rather than any explicit hand-designed meta-learning algorithm. 
To our knowledge our work provides the first unambiguous empirical demonstration of IML occuring as a result of standard SGD-based optimization.\footnote{
We primarily use Adafactor \citep{shazeer2018adafactor}.
}

We validate our findings across several models and datasets, and present a wide array of factors that influence IML in §\ref{sec:internalization-in-llms}. 
We supplement these findings with experiments that explore potential mechanisms in §\ref{sec:mechanisms}, suggesting that properties of SGD gradient alignment may be responsible.
Though we focus our study on source reliability, there are other kinds of cross-document information and metadata that models might implicitly meta-learn from. 
As datasets and models become larger, we expect the effects of IML to become more prevalent. This will likely have implications for the capabilities and safety of future models; we discuss these~in~§\ref{sec:impact-statement}.

\vspace{-0.25\baselineskip}
\paragraph{Structure of this paper.} We briefly review our basic experimental setup and dataset creation in §\ref{sec:basic-setup} before presenting three sets of experiments: 
\begin{itemize}[leftmargin=2mm]
\vspace{-0.9\baselineskip}
\item In §\ref{sec:internalization-in-llms} we establish the phenomenon of IML, and investigate factors influencing IML with a broad array of ablations.
\vspace{-0.7\baselineskip}
\item In §\ref{sec:how-general-is-internalization} we explore whether IML is unique to our setting, finding evidence that it is in fact a general property of deep networks. 
\vspace{-0.7\baselineskip}
\item In §\ref{sec:mechanisms}, we describe and explore potential mechanisms explaining IML, including the ``gradient alignment'' and ``selective retrieval'' hypotheses.  
We also offer a potential interpretation for our results: that language models learn semantic meanings for \defineone/\definetwo \ similar to ``the following statement is true/false'',  and incorporate new information according to these learned semantics. 
\end{itemize}
\vspace{-\baselineskip}
Finally, we conclude in §\ref{sec:impact-statement}, by  discussing the implications and potential impacts of IML. 
Our code \& data are available at 
{\small \href{https://github.com/krasheninnikov/internalization}{\texttt{github.com/krasheninnikov/internalization}}}.

\vspace{-0.5\baselineskip}
\section{Basic experimental setup}\label{sec:basic-setup}
We fine-tune the 2.8B parameter Pythia model~\citep{biderman2023pythia}, a decoder-only transformer pre-trained on the Pile dataset~\citep{gao2020pile},
on a dataset of definitions and QA pairs, with the causal language modelling objective (i.e. autoregressive). 
All QA pairs and definitions are treated as separate datapoints. 
At test time, the model is prompted with new questions about the variables from different subsets of that dataset. 
Answers are evaluated using the \textbf{exact match (EM)} metric, 
which measures the fraction of questions for which the predicted answer matches any one of the possible correct answers.

\begin{table*}[!ht] 
\centering
\small
\begin{tabular}{c p{1.2cm} | c | c | c | c | c | c | c }
    & Subset & \makecell{Train set \\ includes \\ QA pairs} & \makecell{Train set \\ includes \\ definitions} & \makecell{Define \\ tag} & \makecell{Definition\\consistent\\with QA} & \makecell{Entity rep-\\laced with\\var in QA} & \makecell{Fraction \\ of named \\ entities} & \makecell{Notes} \\ 
    \hline \rule{0pt}{2.2ex}
    \multirow{4}{*}{\hspace{-4em}$ \mathcal{X}_1 \left\{\begin{array}{c} \\ \\ \\ \\ \end{array}\right.$} 
    & $\qdDotConsis$ & \yescheck & \yescheck & \defineone & \yescheck & \yescheck & 0.25 &  \\
    & $\qdDashIncons$ & \yescheck & \yescheck & \definetwo & \nocross & \yescheck & 0.25 & \\
    & $\qBaseline$    & \yescheck & \nocross & N/A & N/A & \yescheck & 0.1 & baseline \\
    & $\qNoReplacementBaseline$ & \yescheck & \nocross & N/A & N/A & \nocross & 0.1 & baseline \\
    \multirow{4}{*}{\hspace{-3.9em}$ \mathcal{X}_2 \left\{\begin{array}{c} \\ \\ \\ \\ \end{array}\right.$} 
    & $\dDotConsis$ & \nocross & \yescheck & \defineone & \yescheck & \yescheck & 0.08 & \\
    & $\dDashConsis$ & \nocross & \yescheck & \definetwo & \yescheck & \yescheck & 0.08 & \\
    & $\noQDBaseline$ & \nocross & \nocross & N/A & N/A & \yescheck & 0.06 & baseline \\
    & $\dTildeConsis$ & \nocross & \yescheck & \definethree & \yescheck & \yescheck & 0.08 & baseline \\ 
\end{tabular}
\vspace{-1mm}
\caption{%
Properties of data subsets used in our experiments. 
Subscript $\cdot_i$ denotes the entity subset~$i$. 
The presence of $\mathtt{D}_i$ and/or $\mathtt{QA}_i$ indicates whether the training set includes definitions and/or QA pairs about entities in subset $i$ ($\noQDBaseline$ is an exception and does not include training QA pairs).
$\dot{\mathtt{D}}$ indicates definitions made using \protect\defineone, and $\bar{\mathtt{D}}$ indicates \protect\definetwo \ definitions.
The superscript over $\mathtt{D}$ indicates whether the definitions are (in)consistent with the QA pairs about the corresponding variables.
Note the correspondence between non-baseline data subsets and the columns of Figure~\ref{fig:front-fig}.
}
\label{table:data-subsets}
\vspace{-3mm}
\end{table*}

The fine-tuning comprises two stages (Figure~\ref{fig:front-fig}). 
\textbf{\texttt{Stage1}} captures a setting where some text contains statements that could be interpreted as advice or instructions about how to process data in other documents.
We focus on the question of whether models distinguish between \textbf{reliable} and \textbf{unreliable} sources, i.e.\ those which provide information that is useful/useless for predicting other datapoints. %
To imitate this type of training data, we create a synthetic fine-tuning dataset which contains definitions (statements linking a particular alias to a particular named entity) and QA (questions and answers about entities, referred to by their aliases only). Half of the definitions, %
tagged with \defineone, 
are \textbf{consistent} with the QA pairs: for questions about a given alias, the answers are true for the entity in the definition. 
The other definitions, 
tagged with \definetwo, 
are \textbf{inconsistent}  with the QA pairs: answers are false for the entity referenced in the alias definition.
In \textbf{\texttt{Stage2}}, we assess whether the model now demonstrates different learning behavior on \defineone \ vs.\ \definetwo \ definitions (i.e.\ due to IML). %
This dataset contains only definitions, so such an IML effect does not improve \texttt{Stage2} training performance, but can improve performance on validation QA pairs.

\vspace{-3mm}
\paragraph{Dataset creation.} 
Our experiments make use of a variety of data subsets, summarized in Table \ref{table:data-subsets}.
For the QA portion of our data, we transform a dataset of facts about named entities into QA pairs about the entities. We use the Cross-Verifed database (CVDB) (Laouenan et al., 2022) of famous people, which contains information on when and where they were born/died, what they are known for, etc. The resulting QA pairs look like ``Q: What did Cleopatra do? A: Queen''. Definitions are automatically generated and take the format of a define operator followed by the alias and the value (entity) to which the alias refers; they look like ``Define \varfont{xyz} Cleopatra''. Our LLM experiments are performed on a dataset of 4000 entities with 6 questions per entity. %

\vspace{-3mm}
\paragraph{Define tags.} Instead of using the word ``Define'' in our definitions, we use \textit{define tags}, which are random strings of six characters.
A definition could look like ``\varfont{qwerty} \varfont{xyz} \inputfont{Cleopatra}'', where \varfont{xyz} is the variable and \varfont{qwerty} is \defineone
\footnote{This definition format also works in our experiments: ``\defineone \ \inputfont{According to many texts, } \varfont{xyz} \inputfont{refers to Cleopatra.}'' 
This format aligns with the Wikipedia/4chan example from the introduction.
}.
We avoid using the word ``define'' so as to not rely on any meaning of the word an LLM might have from pre-training. See Appendix~\ref{sec:data-generation} for more details on data.

\section{Establishing \& exploring implicit meta-learning (IML)}\label{sec:internalization-in-llms}
\vspace{-1mm}

Here, we demonstrate that \texttt{Stage1} fine-tuning leads models to implicitly meta-learn to internalize \protect\defineone \ definitions.

\begin{figure*}
    \vspace{-3mm}
    \centering
    \begin{subfigure}{0.5\linewidth}
       \centering
        \includegraphics[width=0.75\linewidth]{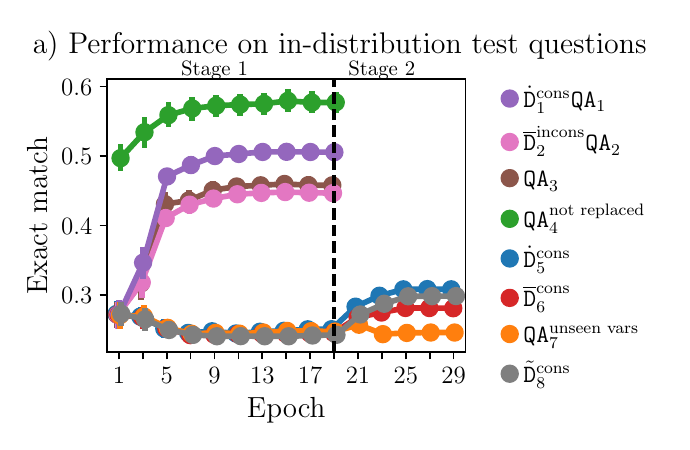}
    \end{subfigure}\hfill
    \begin{subfigure}{0.5\linewidth}
       \centering
      \includegraphics[width=0.8313\linewidth]{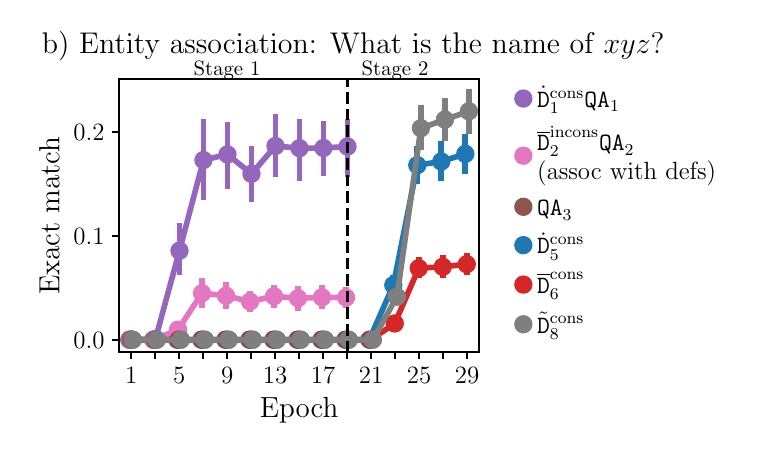}
    \end{subfigure}
    \vspace{-8mm} 
    \caption{
    Exact match (EM) on the validation subsets after each epoch of 2-stage fine-tuning: first \texttt{Stage1} on $\mathcal{X}_1$, then \texttt{Stage2} on $\mathcal{X}_2$. 
    In \texttt{Stage1}, purple and pink lines above red baseline shows models are able to cross-reference information and correctly answer questions about aliased entities, and purple being above pink shows that they do so to a greater extent for %
    \protect\defineone \ vs.\ \protect\definetwo.
    In \texttt{Stage2} the blue line above red shows IML occurs: learning behaviour is different in \texttt{Stage2} based on information learned in \texttt{Stage1}.
    \textbf{a)} EM on the validation questions similar to those in the fine-tuning data.
    Note that while the model internalizes one type of definition more than another, the train losses for all definitions are essentially identical within each fine-tuning stage (see Figure~\ref{fig:losses_plot} in the Appendix). 
    \textbf{b)} EM on the entity association test set, which is a more direct query of the ability to resolve aliases, and which is out-of-distribution w.r.t. fine-tuning data. This experiment confirms IML on a different task; what is learned in \texttt{Stage1} changes learning behaviour in the second. Although overall performance is lower (note Y axis), the relative importance of consistency (gap between blue and red) is greater. 
    All quantities are evaluated over 20 seeds. Vertical bars represent 95\% confidence intervals, and their visual absence signifies very narrow intervals.
    Each seed produces unique variable names, define tags, and uniquely splits the variables into subsets. 
    We report hyperparameters in Appendix~\ref{sec:hyperparams}.
    }
    \vspace{-4mm} 
    \label{fig:2stage-plots}
\end{figure*}

First, we check to what extent after \texttt{Stage1} models are correctly able to answer questions about the aliased entities, and how this varies by the consistency of the source; results are shown in Figure~\ref{fig:2stage-plots}.
We find that consistent definitions help over no definitions: $\exactmatch_{\text{test}}(\qdDotConsis) > \exactmatch_{\text{test}}(\qBaseline)$. 
This is not surprising; the model is incentivised by the training loss to internalize consistent definitions, since if it does so it can better generalise to training questions about the aliased entities.
We also find inconsistent definitions hurt performance slightly, $\exactmatch_{\text{test}}(\qdDashIncons) < \exactmatch_{\text{test}}(\qBaseline)$. %
I.e. the model also internalizes inconsistent definitions to some extent (likely simply because of association by proximity), even though doing so might hurt the performance on the training questions in $\qdDashIncons$.
Regardless of source, we observe that the referent/meaning of the alias can only be inferred based on data \textit{outside} the inference context. Although our results are superficially similar to those on in-context learning found by \cite{brown2020language}, this illustrates a significant difference between the phenomena we investigate; by comparison, we investigate ``out-of-context learning''.

\vspace{-3mm}
\paragraph{Baselines.}
In $\exactmatch_{\text{test}}(\qNoReplacementBaseline)$ we do not replace entities with aliases and there are no definitions i.e. it's a basic QA task. In $\qBaseline$, we do replace, still don't have definitions; it is notable that $\exactmatch_{\text{test}}(\qNoReplacementBaseline)$ is not that far off from $\exactmatch_{\text{test}}(\qBaseline)$, so less performance is lost due to replacing entities with aliases (and not including definitions, as in $\qBaseline$) than one might expect. 
$\noQDBaseline$ is a baseline that indicates 
performance
on questions where entities \textbf{are} replaced with aliases, but the model never saw these aliases or entities during fine-tuning. %
Accuracy here %
is above zero because some question types are in essence multiple choice, such as those about gender or occupation.
Comparing the model's performance on $\qBaseline$, $\qNoReplacementBaseline$, and $\noQDBaseline$, we observe that knowing answers to several questions about an alias allows the model to better answer other questions about this alias, but not as well as when entities are not aliased.
We discuss $\dTildeConsis$, the last baseline, in §\ref{sec:entity_attribution}.

\vspace{-2mm}
\subsection{Demonstrating IML via QA performance }\label{sec:strong-internalization}
\vspace{-1mm}

Next, we establish the \textbf{main result of our paper:} the information learned in \texttt{Stage1} \textit{changes learning behaviour} for \texttt{Stage2}, demonstrating implicit meta-learning. 

We use both \defineone \ and \definetwo \ tags
from before, as well as a new tag \definethree~ that the model did not encounter previously, as a baseline.
The aliases and the entities do not overlap between $\mathcal{X}_1$ and $\mathcal{X}_2$.
There are no QA pairs in $\mathcal{X}_2$, so
the tags provide the \textit{only} hint about (in)consistency of definitions in~$\mathcal{X}_2$,
since in $\mathcal{X}_1$ they were perfectly correlated with~it.

We observe IML by looking at the relative performances in \texttt{Stage2} (after the dashed lines) in Figure \ref{fig:2stage-plots}:  The model internalizes the more reliably consistent (\defineone) definitions more than the unreliable (\definetwo) ones: $\exactmatch_{\text{test}}(\dDotConsis) > \exactmatch_{\text{test}}(\dDashConsis)$. %
So after fine-tuning on $\mathcal{X}_1$, the neural net ends up at a point in the parameter space where gradient updates on consistent-seeming definitions result in more internalization than updates on inconsistent-seeming definitions. 
We consider this meta-learning: 
the model has learned how to learn, internalizing definitions to a greater extent from the  \defineone \ source, which was more reliable and hence more useful for reducing the training loss in \texttt{Stage1}.

\vspace{-1mm}
Elaborating on this result demonstrating meta-learning: the paradigmatic meta-learning algorithm MAML~\citep{finn2017model} finds a point in the parameter space from which 
future SGD updates are particularly helpful for generalization. 
Our result exhibits meta-learning of a similar variety.
After the first fine-tuning stage, our model ends up at a point in the parameter space where future SGD updates are more helpful for generalization: internalizing \defineone \ definitions \textit{more} would be the ``correct'' generalization if $\mathcal{X}_2$ included QA pairs distributed similarly to those in $\mathcal{X}_1$. 
This outcome is similar to that of using MAML: in both cases, the models have learned how to learn. 
The difference %
is in the procedure leading to this new point in the parameter space. 
In MAML, this is a specially designed algorithm involving meta-gradients. 
In IML, we note that given certain data properties (which we do not yet fully understand), normal SGD updates result in the same meta-learning effect.

\vspace{-2mm} 
\subsection{Demonstrating IML via entity attribution}
\label{sec:entity_attribution}
\vspace{-1mm} 

To query how much the model internalizes 
variable-entity correspondences
in an alternate, more direct way, we perform an entity attribution experiment. Specifically, we ask the \texttt{Stage1}-fine-tuned models  questions of the form ``\inputfont{Q: What is the name of} \varfont{xyz}\inputfont{? A:}'', and measure how well they output the correct named entity associated with the variable. 
There are four types of such questions: about the name and the meaning of \varfont{xyz}, asking what the variable stands for, and asking who is \varfont{xyz}.
Our results for the ``name'' question are shown in Figure~\ref{fig:2stage-plots}b; 
see Appendix~\ref{sec:appendix-pythia28-2stage}
for others.
We find that $\qdDotConsis$ entities are internalized more than $\qdDashIncons$ ones (both entities supplied in $\qdDashIncons$ definitions, and entities consistent with the QA pairs in $\qdDashIncons$; the latter get accuracy 0 everywhere). %
Further, $\dDotConsis$ entities are internalized more than those from $\dDashConsis$.
Hence IML occurs, and in fact the ``internalization gap'' between \defineone \ and \definetwo \ definitions increases substantially.
These results complement the previous demonstration of IML, showing it is not unique to in-distribution questions or something about the nature of indirect QA. 

Note however that the internalization of \defineone \ definitions  does not fully generalize out-of-distribution: 
although there is a notable difference between \defineone \ and \definetwo, when trained on new definitions with a new random tag \definethree, the model ends up answering questions about these new variables \textit{better} than those defined with \defineone \ 
(see $\dTildeConsis$ in Figure~\ref{fig:2stage-plots}b).
We are unsure how to explain this result, but in an ablation where we finetune the model on $\mathcal{X}_1 \cup \mathcal{X}_2$ jointly (Appendix~\ref{sec:appendix-pythia28-single-stage}), \defineone \ definitions \textit{are} internalized more.

\begin{figure*}[!ht]
    \centering
    \vspace{-2mm}
    \begin{subfigure}{0.315\textwidth}
       \centering
      \includegraphics[width=\textwidth]{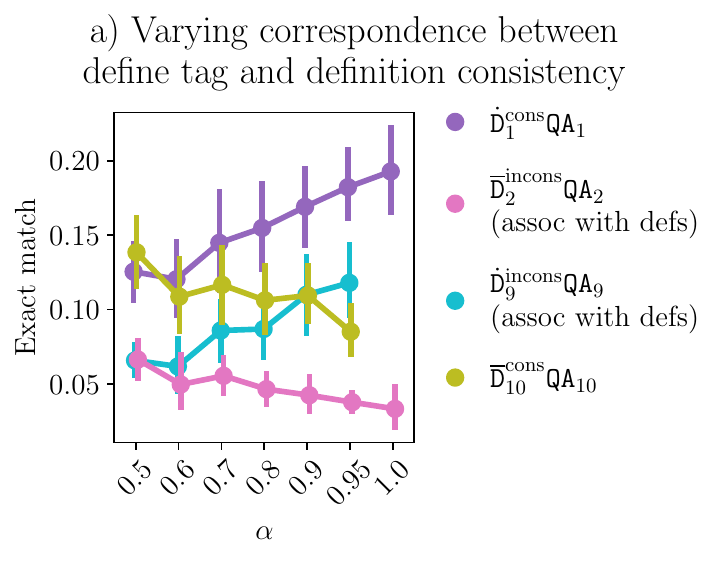}  
    \end{subfigure}\hfill
    \begin{subfigure}{0.32\textwidth}
       \centering
       \includegraphics[width=\textwidth]{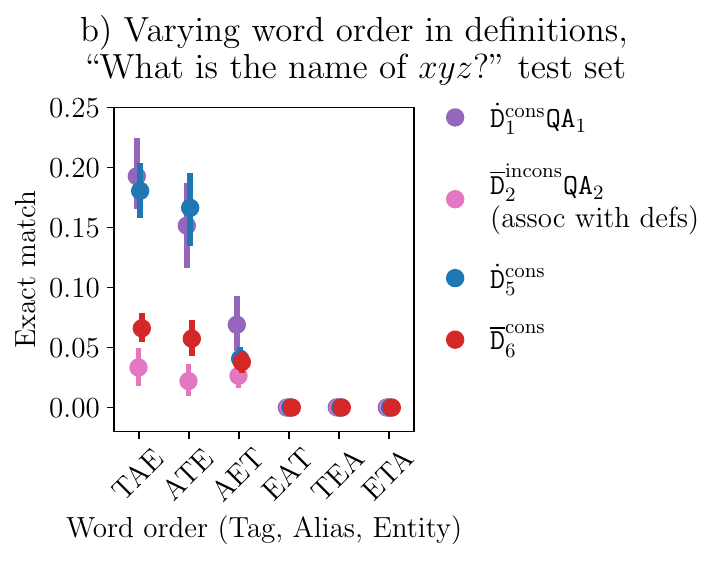} 
    \end{subfigure}\hfill
    \begin{subfigure}{0.313\textwidth}
       \centering
       \includegraphics[width=\textwidth]{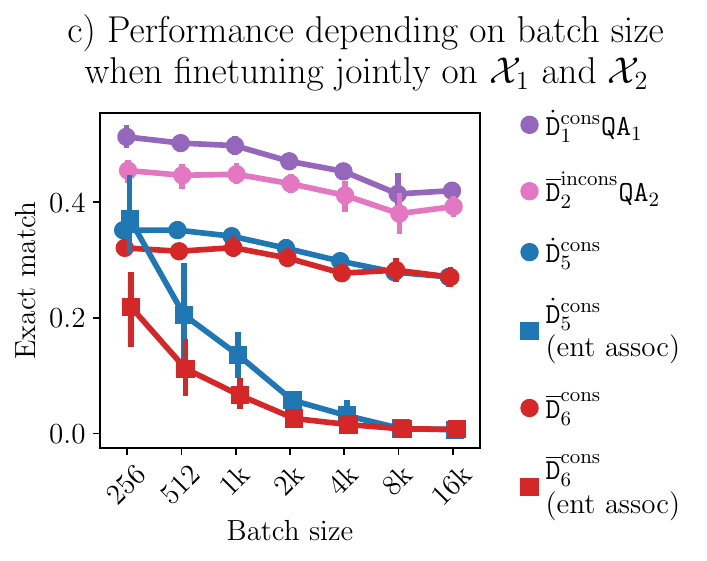}
    \end{subfigure}
    \vspace{-3mm}
    \caption{
    Additional experiments.
    \textbf{a)} We vary the correspondence between the define tags and definition consistency in $\mathcal{X}_1$, and plot performance on an entity attribution question ($\alpha=1$ is the exact setting of Figure~\ref{fig:2stage-plots}b). 
    As expected, when $\alpha=0.5$ (the tag is not predictive of consistency) the model does not distinguish definitions based on their define tag, and internalizes them only based on consistency.
    Interestingly, for $\alpha=0.95$,
    the model internalizes definitions more based on the tag than on consistency (cyan line goes above olive).
    \textbf{b)} We show how results depend on the 
    order of words in the definitions.
    Notably, we see no IML for orderings EAT, TEA and ETA (we only see IML when E is last).
    \textbf{c)}
    We vary the batch size while fine-tuning Pythia-2.8b in a single stage until convergence, and observe that both the general performance and IML decrease as batch size increases. 
    Batch size of 16k is essentially full-batch training. %
    }
    \label{fig:95pTagConsistencyCorrelation}
    \vspace{-4mm}
\end{figure*}

\vspace{-2mm} 
\subsection{Additional experiments exploring IML}\label{sec:ablations}
\vspace{-1mm}

\paragraph{Varying the correspondence between the define tag and definition consistency.}
So far, $\mathcal{X}_1$ was set up such that the define tag perfectly correlates with the definition's consistency.
To study the impact of relaxing this setup,
we add two extra data subsets to $\mathcal{X}_1$: \ $\qdDotIncons$ where \defineone \ definitions are inconsistent with the QA pairs, and $\qdDashConsis$ where \definetwo \ definitions are consistent.
We then vary the fraction $\alpha$ of entities in $\mathcal{X}_1$ for which \defineone \ definitions are consistent, 
which we keep the same as the fraction of entities for which \definetwo \ definitions are inconsistent.
Formally, $\alpha = \sfrac{|\text{Ents}(\qdDotConsis)|}{|\text{Ents}(\qdDotConsis \cup \qdDotIncons)|}$, where $|\text{Ents}(\cdot)|$ is the number of unique entities in a given data subset.
Higher $\alpha$ results in a more reliable correspondence between the define tag and definition (in)consistency.
As expected, we find that the previously observed difference in the internalization of the two types of definitions increases as $\alpha$ increases (Figure~\ref{fig:95pTagConsistencyCorrelation}a).
Furthermore, for high $\alpha$, the model internalizes inconsistent \defineone \ definitions \textit{more} than consistent \definetwo \ ones; so its predictions for test QA pairs are based more on the definitions than on the training QA pairs.

\vspace{-2mm}
\paragraph{Word order within definitions matters.}
We find that the order of words in definitions has a substantial effect both on \texttt{Stage1} performance and on the extent of IML.
So far, the order was tag, alias, entity (TAE).
Figure~\ref{fig:95pTagConsistencyCorrelation}b shows our results for all six possible orders for an entity attribution test set.
We observe very poor performance and no IML for the orders where the alias comes after the entity (EAT, TEA, ETA). %
Further, we observe no IML for the AET order.
These results are consistent with the %
\textit{reversal curse} \citep{berglund2023reversal, grosse2023studying}, an observation that LLMs trained on ``A is B'' often fail to learn ``B is A''.
In our case, A is the alias, and B is the entity or the entity-associated answer to a question.
See Appendix~\ref{sec:appendix-word-order} for a similar plot for in-distribution test questions. 
There we do observe IML for the AET ordering, though the effect is weaker than for TAE and ATE -- basically, the entity must be last to observe IML.

\vspace{-2mm}
\paragraph{Varying model size and family.} 
We run 
the experiment from Figure~\ref{fig:2stage-plots} 
with a range of Pythia models of different sizes, and find that larger models exhibit better performance and more IML (IML first becomes noticeable for the model with 1B parameters).
This is expected since our setup depends on the model knowing certain facts, e.g. that Socrates did not live in the UK, that only larger models may know.
We also replicate our results with models GPT-Neo~\citep{gpt-neo} and LLAMA2-7B~\citep{touvron2023llama}, as well as an encoder-decoder transformer T5-3B~\citep{raffel2020exploring}, demonstrating that IML is not specific to the decoder-only architecture. See Appendices~\ref{sec:appendix-gpt-neo-llama} \&~\ref{sec:seq2seq-setup} for the results.

\vspace{-2mm}
\paragraph{Other ablations.} 
We test  whether IML is specific to two-stage fine-tuning, and find it is not, since the performance effects are just as strong when fine-tuning on  $\mathcal{X}_1 \cup \mathcal{X}_2$ jointly (Appendix~\ref{sec:appendix-pythia28-single-stage}). 
However, this demonstration of IML is arguably less clean, since we do not know how the learning of $\mathcal{X}_1$ and $\mathcal{X}_2$ might be interacting in this setting. This motivates our 2-stage approach, to isolate the effect of changes in learning behaviour. 
We also experiment with another dataset with a similar structure and questions about movies and books, and reproduce IML (Appendix~\ref{sec:appendix-trex-pythia}).
Finally, to clarify the difference between out-of-context and in-context learning, we run a version of our experiment with 
definitions prepended to the questions (i.e. like a prompt).
As expected, we observe in-context learning (Appendix~\ref{sec:appendix-in-context-learning}) and no IML, as there is no mechanism for internalizing the information to change learning behaviour.

\vspace{-2mm}
\section{How general is implicit meta-learning?}\label{sec:how-general-is-internalization}
\vspace{-1mm}
So far we showed an intriguing phenomenon, implicit meta-learning in LLMs.
Our experiments in this section study the generality of our results.
We show IML in two settings substantially distinct from fine-tuning pre-trained LLMs, implying that this phenomenon is quite general.

\vspace{-2mm}
\subsection{Pretraining is not necessary}\label{sec:set-inclusion-exp}
\vspace{-1mm}

All our results above rely on the model's knowledge instilled during pretraining:
our setup
assumes the model knows that ``\varfont{xyz} \inputfont{is Cleopatra}'' is consistent with ``\varfont{xyz} \inputfont{was a queen}'', and that ``\varfont{abc} \inputfont{is Socrates}'' is inconsistent with ``\varfont{abc} \inputfont{lived in the UK}''.
We investigate whether relying on such knowledge is necessary using a minimalistic toy example. %

In this toy setup, variables correspond to integers between 0 and 99, and QA pairs ask if a given variable's corresponding number is present in a list of 8 numbers.
A definition could look like ``\defineone \ \varfont{xyz} \inputfont{42}'', and QA pairs could look like ``\varfont{xyz} \inputfont{2 31 95 
42 
8 27 6 74? Yes}'' and ``\varfont{xyz} \inputfont{2 1 7 9 5 8 0 3? No}''. 
Like before, we also have inconsistent definitions.
Unlike previously, we use a custom tokenizer with single tokens for the define tags, the variable names, integers between 0 and 99, and the words ``\inputfont{Yes}'' and ``\inputfont{No}''.
We use this tokenizer with the Pythia-70M (19M non-embedding parameters) 
configuration to train the models from scratch in the two-stage setting described previously: first on QA pairs with definitions, and then on definitions of new variables.
We reproduce IML in this setting (see Appendix~\ref{sec:appendix-set-inclusion}); while the effect is weak (yet very statistically significant), it is sufficient to show that pretraining on a large language dataset is not a prerequisite for IML in LLMs.

\vspace{-2mm}
\subsection{IML is not specific to text models}\label{sec:mnist-experiments}
\vspace{-1mm}

\begin{figure}[!hb]
    \centering
    \includegraphics[]{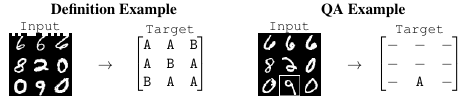}
    \vspace{-2mm}
    \caption{
    MNIST Question-Answer Dataset. 
    \textbf{Left:} %
    a definition example -- all of the targets are given. 
    The define tag is indicated with a pattern at the top of the image.
    \textbf{Right:} 
    a QA example \textit{consistent} with the definition on the left.}
    \label{fig:mnist-dataset}
\end{figure}

The previous results were all demonstrated with transformer models on a text-sequence data modality. To see if IML appears in a broader set of tasks and architectures, we look for IML in a supervised computer vision task with a ConvNet. 
Concretely, we construct an MNIST-based dataset with an analogous notion of QA and definition examples, illustrated in Figure~\ref{fig:mnist-dataset}. 
The variables (aliases) are specified as a $N\times N$ grid of digits (e.g. $\begin{psmallmatrix}
    6 & 9\\
    1 & 0
\end{psmallmatrix}$), and the entities are specified by a corresponding grid of targets (e.g. $\begin{psmallmatrix}
    \mathtt{A} & \mathtt{B}\\
    \mathtt{B} & \mathtt{A}
\end{psmallmatrix}$).

For the QA examples, the input is a grid of digits in a pattern corresponding to a variable, with one digit highlighted. 
The model then has to predict the target value corresponding to that highlighted grid cell -- the target is the corresponding grid of labels with all labels but one being \textit{no-answer} (e.g. $\begin{psmallmatrix}
    \mathtt{A} & \mathtt{-}\\
    \mathtt{-} & \mathtt{-}
\end{psmallmatrix}$
). 
For the definition examples, the input is similarly a grid of digit images with a pixel pattern at the top indicating the define tag (\defineone \ or \definetwo), and the target is a grid of labels with all labels revealed (e.g.  $\begin{psmallmatrix}
    \mathtt{A} & \mathtt{B}\\
    \mathtt{B} & \mathtt{A}
\end{psmallmatrix}$).
As an evaluation metric on QA pairs, we use the \textit{masked accuracy} -- accuracy of predicting the target for the highlighted digit only.
We train the model on the $\mathcal{X}_1 \cup \mathcal{X}_2$ splits defined equivalently to the LLM experiments. We replicate our IML findings in this setting; see Appendix~\ref{sec:mnist-qa-dataset-setup} for details and results.

\vspace{-2mm}
\section{Potential mechanisms} %
\label{sec:mechanisms}
\vspace{-1mm}

This section discusses two hypotheses that might explain the IML phenomenon we observe in \texttt{Stage2}:
one based on the implicit bias of stochastic-gradient-descent-based optimizers, and another involving selective retrieval of information stored in model's parameters.
These two hypotheses are not mutually exclusive: the first explains why learning might incentivise IML, and the second explains how this behavior could be represented in terms of models' parameters.
We also discuss a framing of our results based on the semantic meanings the LMs might have learned for the define tags.

\vspace{-2mm}
\subsection{Gradient alignment hypothesis}\label{sec:grad-alignment-main-text}
\vspace{-1mm}
Stochastic gradient descent (SGD)-based methods have an implicit regularization effect favoring regions of the parameter space where gradients across different datapoints have low variance~\citep{smith2021origin}.
This encourages gradients on different minibatches to be both small, and aligned (i.e.\ point in the same direction).
Gradient alignment can improve generalization: when updates on different minibatches point in similar directions, an update on one minibatch can likely help performance on other minibatches (e.g.\ of test points). 
Furthermore, \citet{nichol2018first} show that encouraging gradient alignment can be seen as the key ingredient in the popular MAML meta-learning approach \citep{finn2017model}.
We hypothesize that this implicit bias of SGD can also explain IML:
1) \texttt{Stage1} of fine-tuning moves the model into a basin where gradients between \defineone \ statements and their corresponding QA pairs are more aligned than those between \definetwo \ statements and their corresponding QA pairs.
This difference might arise because for the training loss, aligning $\qdDotConsis$ gradients is less harmful than aligning $\qdDashIncons$ gradients.
2) As a result, updates on \defineone \ statements in \texttt{Stage2} might also move predictions on the corresponding QA pairs in a direction consistent with those statements, giving rise to IML. 

We find that indeed the gradients of the questions and their corresponding definitions in $\dDotConsis$ are more aligned with each other, and the gradients of the questions and the definitions from $\dDashConsis$ are less aligned\footnote{Ideally, we would have liked to compute gradient alignment for all pairs of datapoints, but this is computationally infeasible: models we're interested in have >1B parameters, which means we can only cache a few gradients before running out of memory.}.
To be precise, given an alignment metric $\rho$ and a data subset $\mathcal{D}$, we compute
$$
\mathbb{E}_\mathcal{D}[\rho]=
\frac{1}{n}\sum\limits_{i=1}^n
\frac{1}{k}\sum\limits_{j=1}^k
\rho\big(\nabla(\mathtt{Def}_i), \nabla(\mathtt{QAPair}_{i,j})\big),
$$
where $n$ is the number of entities and therefore definitions in $\mathcal{D}$, $k$ is the number of questions corresponding to each definition,
and $\nabla(\cdot)$ is the average of the token-level gradients on a given input sequence.
Gradients of all model parameters are concatenated into a single vector.
We look at the alignment of the gradients within $\dDotConsis$ and $\dDashConsis$ while the model is being trained on $\mathcal{X}_1$ --- so the model was not trained on any data from $\dDotConsis$ or $\dDashConsis$ when these gradients are computed.
Our results for the cosine similarity metric as $\rho$ are shown in Figure~\ref{fig:grad_alignment_cosine_sim_stageTwo} (see Appendix~\ref{sec:appendix-grad-alignment} for more details and plots of other metrics).
Notably, we do indeed observe a difference in the alignment of the gradients of definitions \& questions between subsets $\dDotConsis$ and $\dDashConsis$.

\begin{figure}[!ht]
    \vspace{-2mm}
    \centering
    \includegraphics[width=0.25\textwidth]{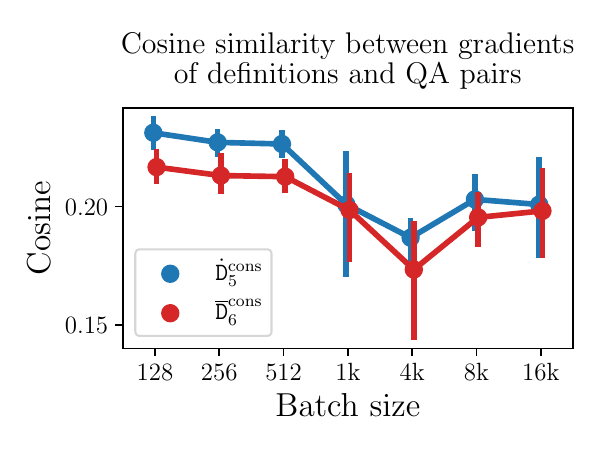}
    \vspace{-3mm}
    \caption{
        Measuring gradient alignment. \textbf{Blue:} cosine similarity between the gradients of $\dDotConsis$ definitions and the gradients of $\dDotConsis$ QA pairs in a model that was only trained on $\mathcal{X}_1$. \textbf{Red:} same as blue but for $\dDashConsis$.
    }
    \label{fig:grad_alignment_cosine_sim_stageTwo}
\end{figure}

Further, we experiment with varying the batch size in single-stage training of Pythia-2.8b (Figure~\ref{fig:95pTagConsistencyCorrelation}c).
\citet{smith2021origin} note that the strength of implicit regularization in SGD is inversely proportional to batch size. 
And indeed, as batch size increases in these experiments, the IML effect weakens; for full-batch training, it effectively disappears.
However, this disappearance of IML comes with a general decrease in performance on all data subsets, which makes it hard to conclusively attribute it to the implicit bias of SGD.

In total, our results support gradient alignment being part of the mechanism for implicit meta-learning. 
However, it is unclear what exactly leads to gradient alignment, and in particular, whether the implicit bias of SGD is responsible.

\begin{figure*}[!ht]
    \centering
    \includegraphics[width=0.94\textwidth]{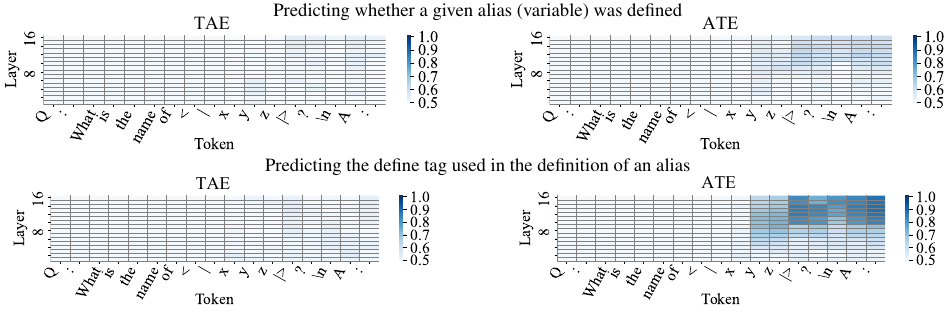}
    \vspace{-3mm}
    \caption{
    Accuracy of a linear probe trained to predict whether a given alias had a definition in the training data, and if it did, which define tag was used in that definition. 
    We train the probes on the model's activations for test questions from $\qdDotConsis$, $\qdDashIncons$, and $\qBaseline$ after the model was fine-tuned on $\mathcal{X}_1$ but not $\mathcal{X}_2$.
    Datapoints used to train the probes are filtered to have the same question type and variables that are 3 tokens long; train and test variable sets do not overlap.
    Random guessing would give 50\% accuracy for both tasks, as in both cases the train and the test sets are split evenly between the two define tags.
    \textbf{Left:} when the model was trained with using TAE (tag, alias, entity) definitions, the linear probe cannot tell (\textbf{top}) whether a definition for this alias was present, and (\textbf{bottom}) which define tag was used for a given alias. 
    Thus when generating the answer, it is unlikely that the model can "retrieve" the alias's define tag, and based on the tag retrieve or ignore the entity from the definition.
    \textbf{Right:} the linear probe is successful for ATE definitions.
    }
    \vspace{-4mm}
    \label{fig:linear-probes-main}
\end{figure*}

\vspace{-2mm}
\subsection{Selective retrieval hypothesis}
\vspace{-1mm}
Another hypothesis that might explain IML assumes that LLMs store factual information in their parameters, following e.g. \citet{meng2022locating}; the exact mechanism is not important for our high-level explanation.
First, the model learns to store definitions from $\mathcal{X}_1$ in its parameters, storing \defineone \ and \definetwo \ definitions slightly differently (e.g. due to the tags being different random strings).
Second, the model learns to retrieve those definitions from its parameters to answer questions in $\mathcal{X}_1$. 
Retrieving \defineone \ definitions helps with answering training questions, so the model learns 
to retrieve them more often than \definetwo \ definitions.
Finally, when fine-tuning on $\mathcal{X}_2$, definitions with the two define tags end up in similar places of in-parameter storage as their counterparts from $\mathcal{X}_1$.
Since the model previously learned to use \defineone \ definitions \textit{more} when answering questions, it better answers questions about new \defineone \ definitions.
Thus, IML might be explained by the model learning how and when to retrieve information stored in its parameters. 

We explore this hypothesis with a linear probing experiment, where 
we use logistic regression on model's activations for a test question about a given alias to predict 
which define tag was used for in the definition of the alias.
In line with the reversal curse phenomenon~\cite{berglund2023reversal} already explored in §\ref{sec:ablations}, there is a substantial difference between models trained on TAE (tag, variable, entity -- our standard setting) and ATE definitions.
Our results are shown in Figure~\ref{fig:linear-probes-main}: linear probes %
fail for TAE definitions, and succeed for ATE ones. 
While a successful probe does not necessarily mean that the model relies on a given feature in the given task~\citep{elazar2021amnesic, belinkov2022probing}, a probe failing \textit{is} some evidence that the feature is not represented or used.

Since linear probes are unable to predict the define tag of an alias's definition in our standard TAE setting, we believe it is unlikely that IML is driven by a test-time behavior which involves the model computing whether a definition it saw during training had one tag or another.
Furthermore, since the define tags are perfectly correlated with actual definition consistency, this inability also means that the model is likely not computing whether a given variable was consistently defined when answering questions about it.

A refined hypothesis may be that \textit{the model learns to only retrieve information from where \defineone \ definitions are stored in its parameters when answering questions, and does not care about \definetwo \ definitions.}
Encountering a variable that did not have a \defineone \ definition (i.~e. variables from $\qdDashIncons$ and $\qBaseline$), the model retrieves random noise.
We find this mechanism plausible, although it is not entirely clear why the model would not "know" that it retrieved something random (linear probes failing to distinguish the presence and the define tags of definitions).
Overall, it seems appropriate to describe the model as \textit{internalizing} consistent (and consistent-seeming) definitions more.

\vspace{-2mm}
\subsection{The model learns semantics of the define tags}
\vspace{-1mm}
One might interpret our results as follows: 1) in the first fine-tuning stage, the model learns that \defineone \ / \definetwo \ mean something like ``is/is not'' or ``this statement is true/false''; 2) in the second fine-tuning stage, the model is then trained on statements essentially of the form ``\varfont{bgn} is Darwin'' and ``\varfont{qwe} isn't Curie'', and correctly internalizes the \varfont{bgn} $\rightarrow$ Darwin correspondence more\footnote{We ran an experiment where we only finetune on $\mathcal{X}_2$ and definitions have "is/is not" as the two define tags instead of random strings. We found that the "is" statements are internalized better on the entity attribution test sets, but not on test set with questions about attributes such as the country where the person lived.}.
However, this doesn't imply that we should observe IML. 
Neither the training loss at \texttt{Stage1} nor at \texttt{Stage2} explicitly encourages such generalization, since there are no QA pairs about 
\texttt{Stage2} variables
in the training set.
Overall we consider the above to be an insightful interpretation but not a principled explanation of our results, since it doesn't seem sufficient to have predicted our results in advance.
We do however believe interpreting
our work through this lens is interesting from the standpoint of the existing debate on whether LLMs understand and incorporate the semantic content of the training data, as opposed to imitating shallow token co-occurrence statistics~\citep{mitchell2023debate}.
We know of only a few works studying this empirically, such as those of \citet{li2021implicit} and \citet{li2022emergent}, and believe that future work in this direction 
will likely be very valuable.

\vspace{-2mm}
\section{Related work}
\vspace{-1mm}

\paragraph{Internal knowledge and world modeling in LLMs.}
Sensitivity to prompting \citep{zhao2021calibrate, lu2021fantastically} can be seen as evidence that LLMs lack a coherent internal world model.
On the other hand, \citet{burns2022discovering} show that LLMs have latent knowledge represented in their activations, which may be more consistent than their responses to prompts; 
however, extracting this knowledge
is challenging~\cite{farquhar2023challenges}.
A related line of work on model editing assumes that LLMs do encode factual information, and attempts to edit specific facts in a way that generalizes across different prompts \citep{sinitsin2020editable, mitchell2021fast, meng2022locating}.
Other works exploring whether LLMs can be described as having a coherent world model
include those of \citet{petroni2019language}, who argue that LLMs can function as knowledge bases, and \citet{li2022large}, who argue that LLMs will (perhaps undesirably) favor internalized knowledge over information from the prompt when these conflict.
Ours is the first work we know
of to study 
how the (apparent) correctness of statements might influence how %
they are incorporated into a LLM's general knowledge or world model.
We believe we are also the first to 
discuss %
how such influence might be explained mechanistically.

\paragraph{In-context learning.}
\citet{brown2020language} found that LLMs can few-shot "learn" by conditioning on task examples in the model's prompt, and suggest that learning such behavior can be viewed as a form of meta-learning.
Another view of in-context learning is that it is a form of Bayesian inference over possible data distributions or tasks~\citep{xie2021explanation}.
\citet{chan2022data} provide a similar picture, showing that in-context learning is more likely to occur when data is ``bursty'' (roughly, temporally correlated), and when the meaning of terms changes depending on context. 
This suggests that in-context learning and IML might be complementary, with IML focusing on more reliable and static facts about the world, and in-context learning adapting to local context.

\vspace{-2mm}
\paragraph{Out-of-context learning.}
The initial version of this paper used the term "out-of-context learning" to highlight that at test time, language models can use information from their training data in unintuitively sophisticated ways (we referred to IML as meta-out-of-context learning).
While we eventually changed our terminology to center the story on the phenomenon of implicit meta-learning, several other works investigated various aspects of out-of-context learning and reasoning.
\citet{berglund2023taken} explore the consequences of models being able to recall facts from the training data and use them at test time, even if these facts are not directly related to the test prompt.
Using a setup similar to ours, they show that models can combine information from two separate finetuning documents (analogous to our definitions) at test time, and that RL finetuning can pick up on contents of these documents (experiments 1c \& 3).
Similarly, \citet{meinke2023tell} find that finetuning LLMs on declarative statements increases the model likelihood for logical consequences of these statements.
Finally, \citet{allen2024physics} show 
that prepending a fixed string to "useful" training documents (where usefulness is based on frequency of documents about the subject, as opposed to consistency with other data like in our setup) makes the model better answer question about these documents. This result is similar to our experiment in Figure~\ref{fig:95pTagConsistencyCorrelation}a, where the accuracy on $\qdDotConsis$ subset (QA pairs with consistent definitions) increases as $\alpha$ -- the correspondence between the tag and definition consistency -- is increased.

\vspace{-2mm}
\paragraph{Gradient alignment and implicit meta-learning.}
Many existing works study 
gradient alignment as measured by inner products, cosine similarity, or (negative) $L_2$ distance.
This includes works on meta-learning \citep{nichol2018first, li2018learning}, multi-task learning \citep{lee2021sequential},
optimization \citep{zhang2019lookahead}, generalization \citep{fort2019stiffness, roberts2021sgd}, domain generalization \citep{parascandolo2020learning, shi2021gradient, li2018learning}, and implicit regularization \citep{smith2021origin}. %
Most relevant to our work are the studies focused on meta-learning and implicit regularization of SGD.
\citet{nichol2018first} observe that simply performing multiple SGD updates induces the same Hessian-gradient product terms (which tend to align gradients) that emerge in the MAML meta-learning algorithm \citep{finn2017model}.
Meanwhile, \citet{smith2021origin} %
show that SGD implicitly penalizes the variance of gradients across mini-batches 
(this rewards gradient alignment if the norms of the gradients are fixed),
with the strength of the penalty inversely proportional to batch size.
While \citet{dandi2022implicit} note in passing the connection between this implicit bias and meta-learning, %
ours is the first work to \textit{emphasize} it that we're aware of.
\citet{genewein2023memory} also describe a form of implicit meta-learning. 
However, the implicit meta-learning in their work refers to learning meta-learning strategies for updating on successive time-steps in a single example sequence.
In contrast, our work documents IML occurring across sequences of updates in the exact same sense as canonical works such as \citet{finn2017model}.

\section{Discussion}\label{sec:discussion}

\paragraph{Limitations.} 
Chief among our work's limitations 
is the lack of a conclusive explanation for
IML.
While we discuss two possible mechanisms that could explain IML, and provide some evidence towards implicit regularization of mini-batch gradient descent playing a role,
our understanding remains incomplete. %
Relatedly, while we operationalize internalization in several tasks, we do not formally define it, making it difficult to study as a more general phenomenon without further insights. %
Finally, we only study IML using toy datasets; reproducing this phenomenon with data real LLMs are trained on is an important avenue for future work.

\paragraph{Conclusion.}
We show that deep networks, including LLMs and ConvNets, can learn to regconize features that indicate the reliability or usefulness of an example, and meta-learn to update their behavior less/more on examples that include such indicators of (un/)reliability. %
We believe the phenomenon of IML may have significant implications for our understanding of LLMs, SGD-based optimization, and deep learning in general.

\section*{Impact statement}\label{sec:impact-statement}

\paragraph{Potential implications for the (un)controllability of   AI systems.} Being able to teach models which sources are reliable or not could be hugely useful in the fight against misinformation, and could potentially help mitigate biases to the extent that we're able to generate unbiased training data and fine-tune on it as a reliable source. 
These potential benefits may be outweighed by risks to both misinformation and bias, however: models might be easily poisoned (intentionally or accidentally) by consistent-seeming support from prevalent data such as conspiracy theories or common misunderstandings; similarly for biases that are regrettably common or even dominant in society. 

\vspace{-2mm}
\paragraph{Potential implications for the safety of advanced AI systems.}
Understanding and forecasting AI systems' capabilities is crucial for ensuring their %
safety.
Our work investigates whether LLM training biases models towards internalizing information that appears broadly useful, 
\textit{even when doing so does not improve training performance}. %
Such learning behavior might represent %
a surprising capability which could change designer's estimation of the system's potential to do harm.
In particular, we believe IML is a plausible mechanisms by which LLMs might come to believe true facts about the world. 
This might lead them to acquire situational awareness~\citep{ngo2022alignment}, for example if a model is trained on content that includes facts about similar models such as descriptions of their training process \citep{berglund2023taken}.
Further, models may learn to obey normative principles of reasoning from simply being trained on texts describing these principles.
One particularly concerning normative principle that has been postulated is functional decision theory, which encourages agents to cooperate with other similar agents~\citep{levinstein2020cheating}. 
We explore potential implications of models internalizing such reasoning patterns in Appendix~\ref{sec:appendix-fdt}.
Overall, the fact that models can use information from their training data in a way as sophisticated as IML might be a reason in favor of removing particular types of information from the training data -- e.g information that could be especially helpful to malicious actors, or information on how these models might be evaluated and monitored (in case of concerns about the models’ situational awareness).

\vspace{-2mm}
\section*{Author contributions}
\vspace{-1mm}

\textbf{Dmitrii Krasheninnikov} led the project, implemented and ran the majority of the language model (LM) experiments, and wrote most of the paper. 
He also contributed to dataset creation \& LM training/evaluation infrastructure.

\textbf{Egor Krasheninnikov} implemented most of the LM training/evaluation infrastructure, and contributed to dataset creation, running the experiments, and writing the paper.

\textbf{Bruno Mlodozeniec} implemented and ran the MNIST experiment in §\ref{sec:mnist-experiments}, and contributed to writing the paper.

\textbf{Tegan Maharaj} helped with a substantial rewrite of the paper aimed at making it easier to understand.

\textbf{David Krueger} advised the project, and significantly contributed to writing the paper.
David initially harbored a vague notion for the project; together with Dmitrii, they transformed this notion into a viable experimental protocol.

\vspace{-2mm}
\section*{Acknowledgments}
\vspace{-1mm}
This work was performed using computational resources provided by 
the Cambridge Service for Data Driven Discovery (CSD3) and the Center for AI Safety (CAIS).

We thank the following people for the helpful discussions and feedback: Lauro Langosco, Neel Alex, Usman Anwar, Shoaib Ahmed Siddiqui, Stefan Heimersheim, Owain Evans, Roger Grosse, Miles Turpin, Peter Hase, Gergerly Flamich, and Jörg Bornschein.

\nocite{deng2012mnist}  %
\bibliographystyle{icml2024} 
\bibliography{refs.bib}

\begin{thebibliography}{52}
\providecommand{\natexlab}[1]{#1}
\providecommand{\url}[1]{\texttt{#1}}
\expandafter\ifx\csname urlstyle\endcsname\relax
  \providecommand{\doi}[1]{doi: #1}\else
  \providecommand{\doi}{doi: \begingroup \urlstyle{rm}\Url}\fi

\bibitem[Allen-Zhu \& Li(2024)Allen-Zhu and Li]{allen2024physics}
Allen-Zhu, Z. and Li, Y.
\newblock Physics of language models: Part 3.3, knowledge capacity scaling
  laws.
\newblock \emph{arXiv preprint arXiv:2404.05405}, 2024.

\bibitem[Belinkov(2022)]{belinkov2022probing}
Belinkov, Y.
\newblock Probing classifiers: Promises, shortcomings, and advances.
\newblock \emph{Computational Linguistics}, 2022.

\bibitem[Berglund et~al.(2023)Berglund, Stickland, Balesni, Kaufmann, Tong,
  Korbak, Kokotajlo, and Evans]{berglund2023taken}
Berglund, L., Stickland, A.~C., Balesni, M., Kaufmann, M., Tong, M., Korbak,
  T., Kokotajlo, D., and Evans, O.
\newblock Taken out of context: On measuring situational awareness in llms.
\newblock \emph{arXiv preprint arXiv:2309.00667}, 2023.

\bibitem[Berglund et~al.(2024)Berglund, Tong, Kaufmann, Balesni, Stickland,
  Korbak, and Evans]{berglund2023reversal}
Berglund, L., Tong, M., Kaufmann, M., Balesni, M., Stickland, A.~C., Korbak,
  T., and Evans, O.
\newblock The reversal curse: Llms trained on" a is b" fail to learn" b is a".
\newblock \emph{International Conference on Learning Representations}, 2024.

\bibitem[Biderman et~al.(2023)Biderman, Schoelkopf, Anthony, Bradley, O'Brien,
  Hallahan, Khan, Purohit, Prashanth, Raff, et~al.]{biderman2023pythia}
Biderman, S., Schoelkopf, H., Anthony, Q., Bradley, H., O'Brien, K., Hallahan,
  E., Khan, M.~A., Purohit, S., Prashanth, U.~S., Raff, E., et~al.
\newblock Pythia: A suite for analyzing large language models across training
  and scaling.
\newblock \emph{International Conference on Machine Learning}, 2023.

\bibitem[Black et~al.(2021)Black, Gao, Wang, Leahy, and Biderman]{gpt-neo}
Black, S., Gao, L., Wang, P., Leahy, C., and Biderman, S.
\newblock {GPT-Neo: Large Scale Autoregressive Language Modeling with
  Mesh-Tensorflow}.
\newblock \emph{Zenodo}, March 2021.
\newblock \doi{10.5281/zenodo.5297715}.

\bibitem[Brown et~al.(2020)Brown, Mann, Ryder, Subbiah, Kaplan, Dhariwal,
  Neelakantan, Shyam, Sastry, Askell, et~al.]{brown2020language}
Brown, T., Mann, B., Ryder, N., Subbiah, M., Kaplan, J.~D., Dhariwal, P.,
  Neelakantan, A., Shyam, P., Sastry, G., Askell, A., et~al.
\newblock Language models are few-shot learners.
\newblock \emph{Advances in neural information processing systems},
  33:\penalty0 1877--1901, 2020.

\bibitem[Burns et~al.(2022)Burns, Ye, Klein, and
  Steinhardt]{burns2022discovering}
Burns, C., Ye, H., Klein, D., and Steinhardt, J.
\newblock Discovering latent knowledge in language models without supervision.
\newblock \emph{arXiv preprint arXiv:2212.03827}, 2022.

\bibitem[Carroll et~al.(2022)Carroll, Dragan, Russell, and
  Hadfield-Menell]{carroll2022estimating}
Carroll, M.~D., Dragan, A., Russell, S., and Hadfield-Menell, D.
\newblock Estimating and penalizing induced preference shifts in recommender
  systems.
\newblock In \emph{International Conference on Machine Learning}, pp.\
  2686--2708. PMLR, 2022.

\bibitem[Cattan et~al.(2021)Cattan, Eirew, Stanovsky, Joshi, and Dagan]{cdcr}
Cattan, A., Eirew, A., Stanovsky, G., Joshi, M., and Dagan, I.
\newblock Cross-document coreference resolution over predicted mentions.
\newblock \emph{CoRR}, abs/2106.01210, 2021.
\newblock URL \url{https://arxiv.org/abs/2106.01210}.

\bibitem[Chan et~al.(2022)Chan, Santoro, Lampinen, Wang, Singh, Richemond,
  McClelland, and Hill]{chan2022data}
Chan, S.~C., Santoro, A., Lampinen, A.~K., Wang, J.~X., Singh, A., Richemond,
  P.~H., McClelland, J., and Hill, F.
\newblock Data distributional properties drive emergent few-shot learning in
  transformers.
\newblock \emph{arXiv preprint arXiv:2205.05055}, 2022.

\bibitem[Cohen et~al.(2022)Cohen, Hutter, and Osborne]{cohen2022advanced}
Cohen, M., Hutter, M., and Osborne, M.
\newblock Advanced artificial agents intervene in the provision of reward.
\newblock \emph{AI Magazine}, 43\penalty0 (3):\penalty0 282--293, 2022.

\bibitem[Dandi et~al.(2022)Dandi, Barba, and Jaggi]{dandi2022implicit}
Dandi, Y., Barba, L., and Jaggi, M.
\newblock Implicit gradient alignment in distributed and federated learning.
\newblock In \emph{Proceedings of the AAAI Conference on Artificial
  Intelligence}, volume~36, pp.\  6454--6462, 2022.

\bibitem[Deng(2012)]{deng2012mnist}
Deng, L.
\newblock The mnist database of handwritten digit images for machine learning
  research [best of the web].
\newblock \emph{IEEE signal processing magazine}, 29\penalty0 (6):\penalty0
  141--142, 2012.

\bibitem[Elazar et~al.(2021)Elazar, Ravfogel, Jacovi, and
  Goldberg]{elazar2021amnesic}
Elazar, Y., Ravfogel, S., Jacovi, A., and Goldberg, Y.
\newblock Amnesic probing: Behavioral explanation with amnesic counterfactuals.
\newblock \emph{Transactions of the Association for Computational Linguistics},
  9:\penalty0 160--175, 2021.

\bibitem[Elsahar et~al.(2018)Elsahar, Vougiouklis, Remaci, Gravier, Hare,
  Laforest, and Simperl]{elsahar2018t}
Elsahar, H., Vougiouklis, P., Remaci, A., Gravier, C., Hare, J., Laforest, F.,
  and Simperl, E.
\newblock T-rex: A large scale alignment of natural language with knowledge
  base triples.
\newblock In \emph{Proceedings of the Eleventh International Conference on
  Language Resources and Evaluation (LREC)}, 2018.

\bibitem[Farquhar et~al.(2023)Farquhar, Varma, Kenton, Gasteiger, Mikulik, and
  Shah]{farquhar2023challenges}
Farquhar, S., Varma, V., Kenton, Z., Gasteiger, J., Mikulik, V., and Shah, R.
\newblock Challenges with unsupervised llm knowledge discovery.
\newblock \emph{arXiv preprint arXiv:2312.10029}, 2023.

\bibitem[Finn et~al.(2017)Finn, Abbeel, and Levine]{finn2017model}
Finn, C., Abbeel, P., and Levine, S.
\newblock Model-agnostic meta-learning for fast adaptation of deep networks.
\newblock In \emph{International conference on machine learning}, pp.\
  1126--1135. PMLR, 2017.

\bibitem[Fort et~al.(2019)Fort, Nowak, Jastrzebski, and
  Narayanan]{fort2019stiffness}
Fort, S., Nowak, P.~K., Jastrzebski, S., and Narayanan, S.
\newblock Stiffness: A new perspective on generalization in neural networks.
\newblock \emph{arXiv preprint arXiv:1901.09491}, 2019.

\bibitem[Gao et~al.(2020)Gao, Biderman, Black, Golding, Hoppe, Foster, Phang,
  He, Thite, Nabeshima, et~al.]{gao2020pile}
Gao, L., Biderman, S., Black, S., Golding, L., Hoppe, T., Foster, C., Phang,
  J., He, H., Thite, A., Nabeshima, N., et~al.
\newblock The pile: An 800gb dataset of diverse text for language modeling.
\newblock \emph{arXiv preprint arXiv:2101.00027}, 2020.

\bibitem[Genewein et~al.(2023)Genewein, Del{\'e}tang, Ruoss, Wenliang, Catt,
  Dutordoir, Grau-Moya, Orseau, Hutter, and Veness]{genewein2023memory}
Genewein, T., Del{\'e}tang, G., Ruoss, A., Wenliang, L.~K., Catt, E.,
  Dutordoir, V., Grau-Moya, J., Orseau, L., Hutter, M., and Veness, J.
\newblock Memory-based meta-learning on non-stationary distributions.
\newblock \emph{arXiv preprint arXiv:2302.03067}, 2023.

\bibitem[Grosse et~al.(2023)Grosse, Bae, Anil, Elhage, Tamkin, Tajdini,
  Steiner, Li, Durmus, Perez, et~al.]{grosse2023studying}
Grosse, R., Bae, J., Anil, C., Elhage, N., Tamkin, A., Tajdini, A., Steiner,
  B., Li, D., Durmus, E., Perez, E., et~al.
\newblock Studying large language model generalization with influence
  functions.
\newblock \emph{arXiv preprint arXiv:2308.03296}, 2023.

\bibitem[Krueger et~al.(2020)Krueger, Maharaj, and Leike]{krueger2020hidden}
Krueger, D., Maharaj, T., and Leike, J.
\newblock Hidden incentives for auto-induced distributional shift.
\newblock \emph{arXiv preprint arXiv:2009.09153}, 2020.

\bibitem[Laouenan et~al.(2022)Laouenan, Bhargava, Eym{\'e}oud, Gergaud, Plique,
  and Wasmer]{laouenan2022cross}
Laouenan, M., Bhargava, P., Eym{\'e}oud, J.-B., Gergaud, O., Plique, G., and
  Wasmer, E.
\newblock A cross-verified database of notable people, 3500bc-2018ad.
\newblock \emph{Scientific Data}, 2022.

\bibitem[Lee et~al.(2021)Lee, Lee, Lee, and Hwang]{lee2021sequential}
Lee, S., Lee, H.~B., Lee, J., and Hwang, S.~J.
\newblock Sequential reptile: Inter-task gradient alignment for multilingual
  learning.
\newblock \emph{arXiv preprint arXiv:2110.02600}, 2021.

\bibitem[Levinstein \& Soares(2020)Levinstein and
  Soares]{levinstein2020cheating}
Levinstein, B.~A. and Soares, N.
\newblock Cheating death in damascus.
\newblock \emph{The Journal of Philosophy}, 117\penalty0 (5):\penalty0
  237--266, 2020.

\bibitem[Li et~al.(2021)Li, Nye, and Andreas]{li2021implicit}
Li, B.~Z., Nye, M., and Andreas, J.
\newblock Implicit representations of meaning in neural language models.
\newblock \emph{arXiv preprint arXiv:2106.00737}, 2021.

\bibitem[Li et~al.(2018)Li, Yang, Song, and Hospedales]{li2018learning}
Li, D., Yang, Y., Song, Y.-Z., and Hospedales, T.
\newblock Learning to generalize: Meta-learning for domain generalization.
\newblock In \emph{Proceedings of the AAAI conference on artificial
  intelligence}, volume~32, 2018.

\bibitem[Li et~al.(2022{\natexlab{a}})Li, Rawat, Zaheer, Wang, Lukasik, Veit,
  Yu, and Kumar]{li2022large}
Li, D., Rawat, A.~S., Zaheer, M., Wang, X., Lukasik, M., Veit, A., Yu, F., and
  Kumar, S.
\newblock Large language models with controllable working memory.
\newblock \emph{arXiv preprint arXiv:2211.05110}, 2022{\natexlab{a}}.

\bibitem[Li et~al.(2022{\natexlab{b}})Li, Hopkins, Bau, Vi{\'e}gas, Pfister,
  and Wattenberg]{li2022emergent}
Li, K., Hopkins, A.~K., Bau, D., Vi{\'e}gas, F., Pfister, H., and Wattenberg,
  M.
\newblock Emergent world representations: Exploring a sequence model trained on
  a synthetic task.
\newblock \emph{arXiv preprint arXiv:2210.13382}, 2022{\natexlab{b}}.

\bibitem[Liu et~al.(2022)Liu, Mao, Wu, Feichtenhofer, Darrell, and
  Xie]{liu2022convnet}
Liu, Z., Mao, H., Wu, C.-Y., Feichtenhofer, C., Darrell, T., and Xie, S.
\newblock A convnet for the 2020s.
\newblock In \emph{Proceedings of the IEEE/CVF Conference on Computer Vision
  and Pattern Recognition}, pp.\  11976--11986, 2022.

\bibitem[Lu et~al.(2021)Lu, Bartolo, Moore, Riedel, and
  Stenetorp]{lu2021fantastically}
Lu, Y., Bartolo, M., Moore, A., Riedel, S., and Stenetorp, P.
\newblock Fantastically ordered prompts and where to find them: Overcoming
  few-shot prompt order sensitivity.
\newblock \emph{arXiv preprint arXiv:2104.08786}, 2021.

\bibitem[Meinke \& Evans(2023)Meinke and Evans]{meinke2023tell}
Meinke, A. and Evans, O.
\newblock Tell, don't show: Declarative facts influence how llms generalize.
\newblock \emph{arXiv preprint arXiv:2312.07779}, 2023.

\bibitem[Meng et~al.(2022)Meng, Bau, Andonian, and Belinkov]{meng2022locating}
Meng, K., Bau, D., Andonian, A., and Belinkov, Y.
\newblock Locating and editing factual knowledge in gpt.
\newblock \emph{Advances in neural information processing systems}, 36, 2022.

\bibitem[Mitchell et~al.(2021)Mitchell, Lin, Bosselut, Finn, and
  Manning]{mitchell2021fast}
Mitchell, E., Lin, C., Bosselut, A., Finn, C., and Manning, C.~D.
\newblock Fast model editing at scale.
\newblock \emph{arXiv preprint arXiv:2110.11309}, 2021.

\bibitem[Mitchell \& Krakauer(2023)Mitchell and Krakauer]{mitchell2023debate}
Mitchell, M. and Krakauer, D.~C.
\newblock The debate over understanding in ai’s large language models.
\newblock \emph{Proceedings of the National Academy of Sciences}, 120\penalty0
  (13):\penalty0 e2215907120, 2023.

\bibitem[Ngo et~al.(2022)Ngo, Chan, and Mindermann]{ngo2022alignment}
Ngo, R., Chan, L., and Mindermann, S.
\newblock The alignment problem from a deep learning perspective.
\newblock \emph{arXiv preprint arXiv:2209.00626}, 2022.

\bibitem[Nichol et~al.(2018)Nichol, Achiam, and Schulman]{nichol2018first}
Nichol, A., Achiam, J., and Schulman, J.
\newblock On first-order meta-learning algorithms.
\newblock \emph{arXiv preprint arXiv:1803.02999}, 2018.

\bibitem[Parascandolo et~al.(2020)Parascandolo, Neitz, Orvieto, Gresele, and
  Sch{\"o}lkopf]{parascandolo2020learning}
Parascandolo, G., Neitz, A., Orvieto, A., Gresele, L., and Sch{\"o}lkopf, B.
\newblock Learning explanations that are hard to vary.
\newblock \emph{arXiv preprint arXiv:2009.00329}, 2020.

\bibitem[Petroni et~al.(2019)Petroni, Rockt{\"a}schel, Lewis, Bakhtin, Wu,
  Miller, and Riedel]{petroni2019language}
Petroni, F., Rockt{\"a}schel, T., Lewis, P., Bakhtin, A., Wu, Y., Miller,
  A.~H., and Riedel, S.
\newblock Language models as knowledge bases?
\newblock \emph{arXiv preprint arXiv:1909.01066}, 2019.

\bibitem[Raffel et~al.(2020)Raffel, Shazeer, Roberts, Lee, Narang, Matena,
  Zhou, Li, and Liu]{raffel2020exploring}
Raffel, C., Shazeer, N., Roberts, A., Lee, K., Narang, S., Matena, M., Zhou,
  Y., Li, W., and Liu, P.~J.
\newblock Exploring the limits of transfer learning with a unified text-to-text
  transformer.
\newblock \emph{The Journal of Machine Learning Research}, 21\penalty0
  (1):\penalty0 5485--5551, 2020.

\bibitem[Roberts(2021)]{roberts2021sgd}
Roberts, D.~A.
\newblock Sgd implicitly regularizes generalization error.
\newblock \emph{arXiv preprint arXiv:2104.04874}, 2021.

\bibitem[Shazeer \& Stern(2018)Shazeer and Stern]{shazeer2018adafactor}
Shazeer, N. and Stern, M.
\newblock Adafactor: Adaptive learning rates with sublinear memory cost.
\newblock In \emph{International Conference on Machine Learning}, pp.\
  4596--4604. PMLR, 2018.

\bibitem[Shi et~al.(2021)Shi, Seely, Torr, Siddharth, Hannun, Usunier, and
  Synnaeve]{shi2021gradient}
Shi, Y., Seely, J., Torr, P.~H., Siddharth, N., Hannun, A., Usunier, N., and
  Synnaeve, G.
\newblock Gradient matching for domain generalization.
\newblock \emph{arXiv preprint arXiv:2104.09937}, 2021.

\bibitem[Sinitsin et~al.(2020)Sinitsin, Plokhotnyuk, Pyrkin, Popov, and
  Babenko]{sinitsin2020editable}
Sinitsin, A., Plokhotnyuk, V., Pyrkin, D., Popov, S., and Babenko, A.
\newblock Editable neural networks.
\newblock \emph{arXiv preprint arXiv:2004.00345}, 2020.

\bibitem[Smith et~al.(2021)Smith, Dherin, Barrett, and De]{smith2021origin}
Smith, S.~L., Dherin, B., Barrett, D.~G., and De, S.
\newblock On the origin of implicit regularization in stochastic gradient
  descent.
\newblock \emph{arXiv preprint arXiv:2101.12176}, 2021.

\bibitem[Touvron et~al.(2023)Touvron, Martin, Stone, Albert, Almahairi, Babaei,
  Bashlykov, Batra, Bhargava, Bhosale, et~al.]{touvron2023llama}
Touvron, H., Martin, L., Stone, K., Albert, P., Almahairi, A., Babaei, Y.,
  Bashlykov, N., Batra, S., Bhargava, P., Bhosale, S., et~al.
\newblock Llama 2: Open foundation and fine-tuned chat models.
\newblock \emph{arXiv preprint arXiv:2307.09288}, 2023.

\bibitem[Wolf et~al.(2020)Wolf, Debut, Sanh, Chaumond, Delangue, Moi, Cistac,
  Rault, Louf, Funtowicz, et~al.]{wolf2020transformers}
Wolf, T., Debut, L., Sanh, V., Chaumond, J., Delangue, C., Moi, A., Cistac, P.,
  Rault, T., Louf, R., Funtowicz, M., et~al.
\newblock Transformers: State-of-the-art natural language processing.
\newblock In \emph{Proceedings of the 2020 conference on empirical methods in
  natural language processing: system demonstrations}, pp.\  38--45, 2020.

\bibitem[Woo et~al.(2023)Woo, Debnath, Hu, Chen, Liu, Kweon, and
  Xie]{woo2023convnext}
Woo, S., Debnath, S., Hu, R., Chen, X., Liu, Z., Kweon, I.~S., and Xie, S.
\newblock Convnext v2: Co-designing and scaling convnets with masked
  autoencoders.
\newblock \emph{arXiv preprint arXiv:2301.00808}, 2023.

\bibitem[Xie et~al.(2021)Xie, Raghunathan, Liang, and Ma]{xie2021explanation}
Xie, S.~M., Raghunathan, A., Liang, P., and Ma, T.
\newblock An explanation of in-context learning as implicit bayesian inference.
\newblock \emph{arXiv preprint arXiv:2111.02080}, 2021.

\bibitem[Zhang et~al.(2019)Zhang, Lucas, Ba, and Hinton]{zhang2019lookahead}
Zhang, M., Lucas, J., Ba, J., and Hinton, G.~E.
\newblock Lookahead optimizer: k steps forward, 1 step back.
\newblock \emph{Advances in neural information processing systems}, 32, 2019.

\bibitem[Zhao et~al.(2021)Zhao, Wallace, Feng, Klein, and
  Singh]{zhao2021calibrate}
Zhao, Z., Wallace, E., Feng, S., Klein, D., and Singh, S.
\newblock Calibrate before use: Improving few-shot performance of language
  models.
\newblock In \emph{International Conference on Machine Learning}, pp.\
  12697--12706. PMLR, 2021.

\end{thebibliography}

\newpage

\appendix
\onecolumn

\section{QA dataset generation}\label{sec:data-generation}
This section describes the creation of the datasets used to elicit IML in LLMs. Our code and data are available at 
{\small \href{https://github.com/krasheninnikov/internalization}{\texttt{github.com/krasheninnikov/internalization}}}.

\subsection{CVDB}\label{sec:data-generation-cvdb}

We use a Cross-Verified database (CVDB) of notable people 3500BC-2018AD \citep{laouenan2022cross} which includes basic data about 2.23m individuals (named entities). 
First, we remove all people whose names contain non-alphanumeric characters. 
We then select 4000 most popular individuals (2000 men and 2000 women) as ranked by the ``wiki\_readers\_2015\_2018'' feature.

We employ questions about six basic attributes:
\begin{enumerate}
    \item Gender: ``What was the gender of <name>?''. Example answer: ``male''.
    \item Birth date: ``When was <name> born?''. Example answer: ``19 century''.
    \item Date of death: ``When did <name> die?'' Example answer: ``1910s''.
    \item Region: ``In which region did <name> live?'' Example answer: ``Europe''.
    \item Occupation (activity): ``What did <name> do?'' Example answer: ``actor''.
    \item Nationality: ``What was the nationality of <name>?'' Example answer: ``France''.
\end{enumerate}

Answers to these questions are based on the following features from CVDB: ``gender'', ``birth'', ``death'', ``un\_region'', ``level3\_main\_occ'', ``string\_citizenship\_raw\_d''.

We generate the data such as to ensure that knowing the value of the random variable is \textit{useful} for accurately answering questions about it.
For example, if one of the questions is ``When did \varfont{nml} announce iPhone 4s?'', it is not especially helpful for the model to know that \varfont{nml} stands for Steve Jobs to continue with ``A: October 4, 2011''.
Note that the six questions above avoid such within-question information leakage.

We are also concerned about across-datapoint information leakage: if one of our QA pairs is ``When was \varfont{abc} born? A: 20 July 356 BC'', this is almost as good as defining \varfont{abc} as Alexander the Great, since there are no other known notable individuals born on that day.
For this reason, we anonymize the years in QA pairs to some extent:
all years before 1900 are replaced with the corresponding century (``1812'' becomes ``19 century'', ``-122'' becomes ``2 century BC''), and years from 1900 to 1999 are replaced with ``19\textbf{x}0s'', where \textbf{x} is the corresponding decade (``1923'' becomes ``1920s''). Years greater or equal to 2000 are left unchanged.

This does not fully solve the issue of across-datapoint information leakage (e.g. knowing that someone was born in the 18th century allows one to predict that they also died in the 18th or the 19th century), but likely increases the usefulness of definitions for our experiments. Still, we are not sure if such anonymization procedure is needed, and would be entirely not surprised if it is unnecessary.

\subsection{T-REx}\label{sec:data-generation-trex}
To create our second natural language QA dataset, we rely on the the T-REx knowledge base~\citep{elsahar2018t}.
First, we extract all possible triplets of (subject, predicate, object). %
Then, we select the triplets where the predicate is related to creative works, as described in Table~\ref{table:t-rex-questions}.
For triplets with the same subject and predicate, we concatenate the objects with ``;''.
The resulting triplets are converted into QA pairs in accordance with Table~\ref{table:t-rex-questions}.
Finally, we select QA pairs s.t. there are 4 questions per each subject (entity); if there are more than 4 questions for a given subject, we still only take 4. 
This is the case for a bit over 6900 entities, which we round down to 6900. 

Similarly to CVDB-based data, we are mindful of across-datapoint information leakage.
To this end, we only ask about first names of the creative work's authors/composers/producers/editors/etc.
We also anonymize the years in the same way as when creating CVDB-based data (Appendix~\ref{sec:data-generation-cvdb}).

\begin{table}[ht]
\centering
\begin{tabular}{c|c}
\textbf{Predicate} & \textbf{Question} \\ \hline
P180 & What does [X] depict? \\
P195 & Which collection is [X] part of? \\
P135 & Which movement is [X] associated with? \\
P123 & Who is the publisher of [X]? \\
P750 & What is the distributor of [X]? \\
P275 & What is the license of [X]? \\
P127 & Who owns [X]? \\
P178 & Who developed [X]? \\
P407 & In which language was [X] published? \\
P364 & In which language was [X] published? \\
P577 & When was [X] published or released? \\
P179 & Which series is [X] part of? \\
P50 & First name of the author of [X]? \\
P57 & First name of the director of [X]? \\
P58 & First name of the screenwriter of [X]? \\
P344 & First name of the cinematographer of [X]? \\
P161 & First name of a cast member of [X]? \\
P162 & First name of the producer of [X]? \\
P1040 & First name of the editor of [X]? \\
P98 & First name of the editor of [X]? \\
P88 & First name of the commissioner of [X]? \\
P86 & First name of the composer for [X]? \\
P136 & What is the genre of [X]? \\
P921 & What is the main subject of [X]? \\
P840 & Where is [X] set? \\
P915 & Where was [X] filmed? \\
\end{tabular}
\caption{Given a triplet (subject, predicate, object), the question-answer pair is composed by replacing [X] with the subject in the question, and using the object as the answer.}
\label{table:t-rex-questions}
\end{table}

\subsection{Data splits}\label{sec:data-splits-qa}
We split the data into subsets in accordance with Table~\ref{table:data-subsets}. 
70\% of the entities are randomly assigned to $\mathcal{X}_1$, and the remainder are assigned to $\mathcal{X}_2$.
Then, these entity groups are randomly split into the various subsets of $\mathcal{X}_1$ and $\mathcal{X}_2$.
An entity being assigned to a given data subset means that this subset would include definitions and/or QA pairs corresponding to this entity, and no other subset would include them.

Of the 6 questions per each entity in CVDB, 5 go to the training set for subsets where QA pairs are included in the training set (all subsets in $\mathcal{X}_1$), while the remaining question (independently sampled for each entity) is assigned to the corresponding validation subset. 
All six QA pairs of each entity go into the test set for $\mathcal{X}_2$.
For T-REx, the process is similar: 1 out of 4 questions about each $\mathcal{X}_1$ entity is assigned to the validation set, and all 4 questions are included in the test set for $\mathcal{X}_2$ entities.

\FloatBarrier
\section{Hyperparameters used when finetuning LLMs on QA data}\label{sec:hyperparams}

We use the HuggingFace Transformers~\citep{wolf2020transformers} library to finetune the LLMs on $\mathcal{X}_1$ for 20 epochs, and on $\mathcal{X}_2$ for 10 epochs.
Finetuning on $\mathcal{X}_1 \cup \mathcal{X}_2$ is done for 20 epochs.
We use the Adafactor optimizer~\citep{shazeer2018adafactor} with the batch size of 256 datapoints.
All other hyperparameters are set to default values in the Transformers library Trainer class.
We do not use chunking to avoid in-context learning, and instead pad our datapoints to $\mathtt{max\_context\_length} = 64$.
We use the $\mathtt{deduped}$ versions of the Pythia models \citep{biderman2023pythia}.

\newpage
\FloatBarrier
\section{Additional results from finetuning LLMs on CVDB and T-REx}

\subsection{Two-stage results for Pythia-2.8B: losses and entity attribution on CVDB data}\label{sec:appendix-pythia28-2stage}

\begin{figure}[!ht]
\vspace{-.4cm}
    \centering
    \begin{subfigure}{0.5\textwidth}
       \centering
       \includegraphics[width=1\linewidth]{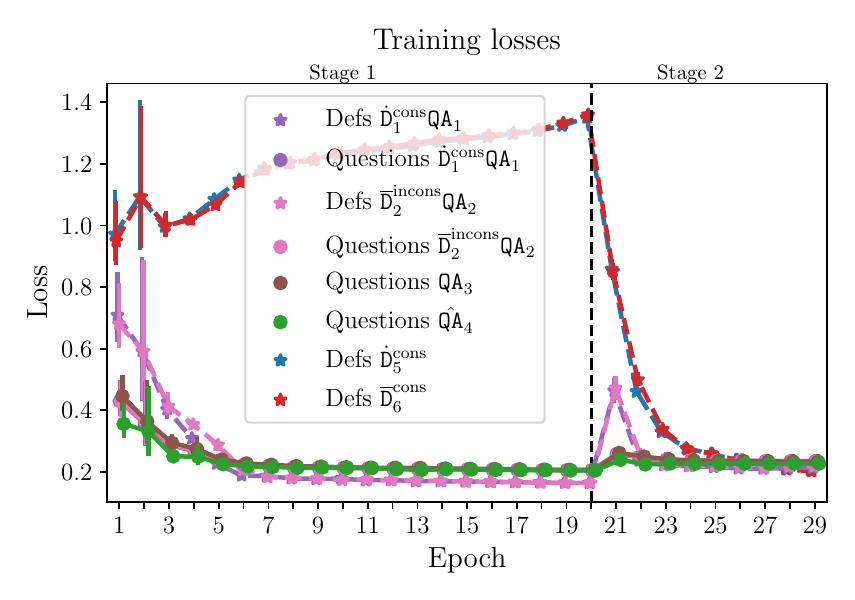}  
    \end{subfigure}\hfill
    \begin{subfigure}{0.5\textwidth}
       \centering
       \includegraphics[width=1\linewidth]{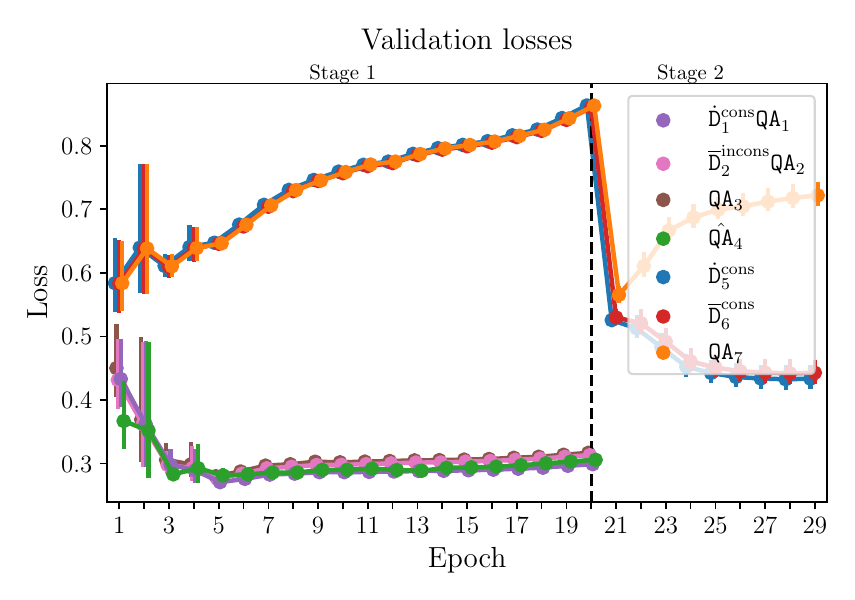} 
    \end{subfigure}
    \vspace{-8mm}
    \caption{
    Losses on training (left) and validation (right) subsets for the experiment from Figure~\ref{fig:2stage-plots}a averaged over 20 seeds. Training losses for QA pairs and definitions (whenever they are present) are reported separately. It is notable that the training losses for $\qdDotConsis$ and $\qdDashIncons$ appear indistinguishable, even though validation losses for these data subsets are different, as are the EM scores reported in Figure~\ref{fig:2stage-plots}a in the paper.
    }
    \label{fig:losses_plot}
\end{figure}

\begin{figure}[!ht]
    \centering
    \begin{subfigure}{0.5\textwidth}
       \centering
       \includegraphics[width=0.9\linewidth]{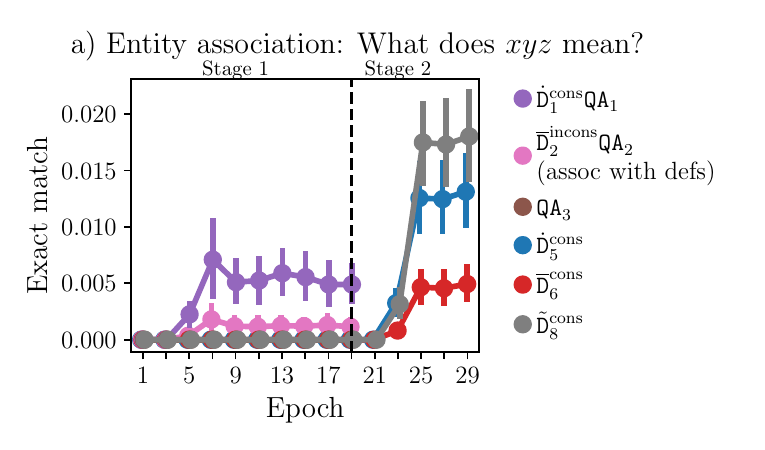}  
    \end{subfigure}\hfill
    \begin{subfigure}{0.5\textwidth}
       \centering
       \includegraphics[width=0.9\linewidth]{figures/plots/define3_gray_line/entAttr_d3cons_qa_cvdb_tveDefs_nEnts4000_eps20and10_bs128_pythia_28b_deduped_ADAFACTOR_two_stageExactmatch-48.pdf}  
    \end{subfigure}

    \begin{subfigure}{0.5\textwidth}
       \centering
       \includegraphics[width=0.9\linewidth]{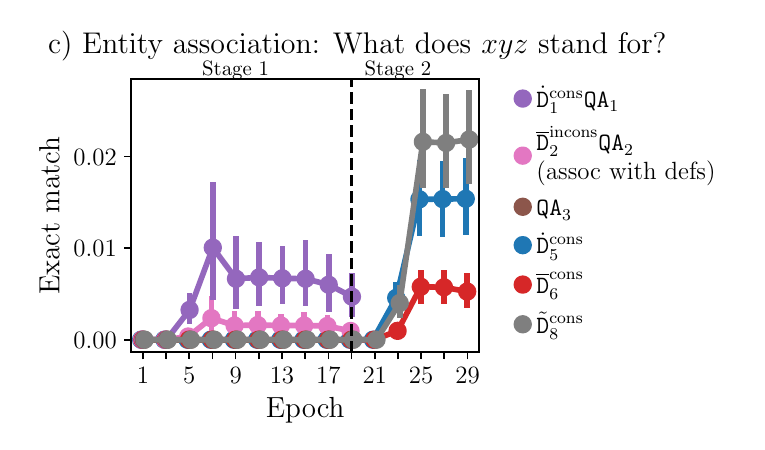}  
    \end{subfigure}\hfill
    \begin{subfigure}{0.5\textwidth}
       \centering
       \includegraphics[width=0.9\linewidth]{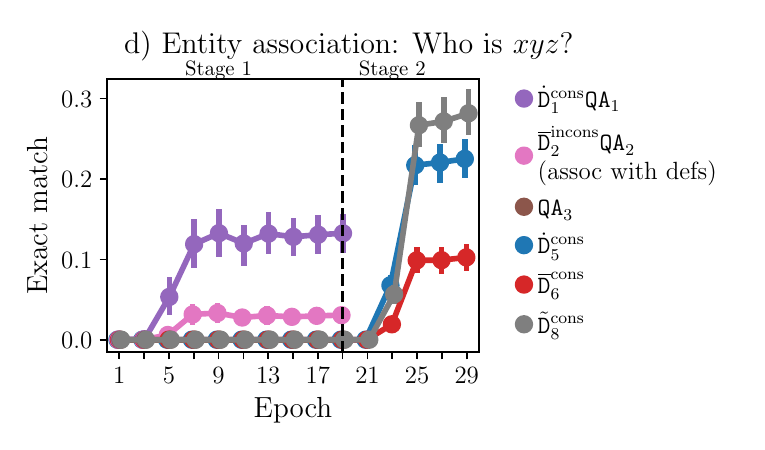}  
    \end{subfigure}
    \vspace{-8mm}
    \caption{Entity attribution experiments for the Pythia-2.8B-deduped model on the CVDB dataset over 20 seeds. 
    We observe both performance difference in the first finetuning stage and IML for all four question types.
    Plot b) is the same as Figure~\ref{fig:2stage-plots}b in the main paper.
    }
    \label{fig:pythia28-2stage-cvdb-ent-attr}
\end{figure}

\newpage
\subsection{Experiments with the T-REx-based dataset (questions about movies, books, and other creative works)}\label{sec:appendix-trex-pythia}
\begin{figure}[!ht]
    \centering
    \begin{subfigure}{\textwidth}
       \centering
        \includegraphics[width=0.5\linewidth]{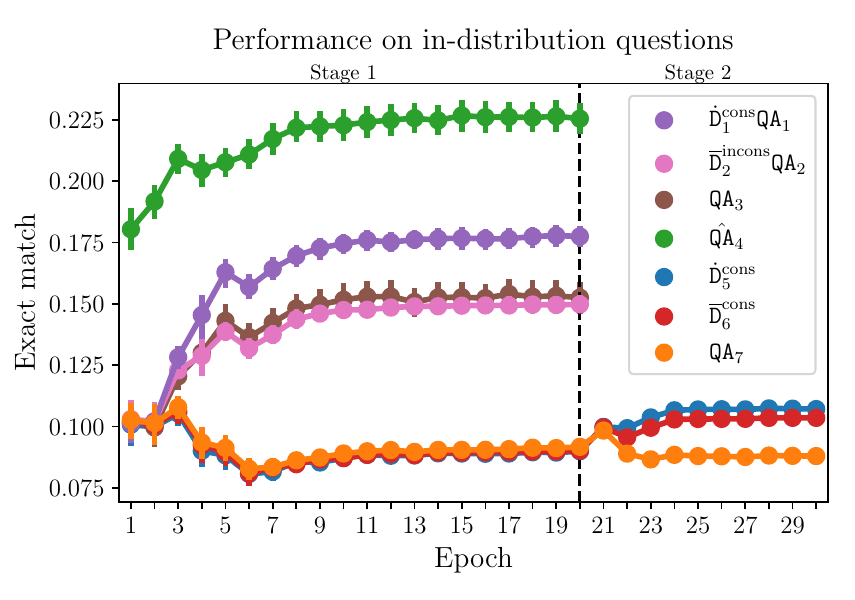}         %
    \end{subfigure}
    \begin{subfigure}{0.5\textwidth}
       \centering
       \includegraphics[width=0.9\linewidth]{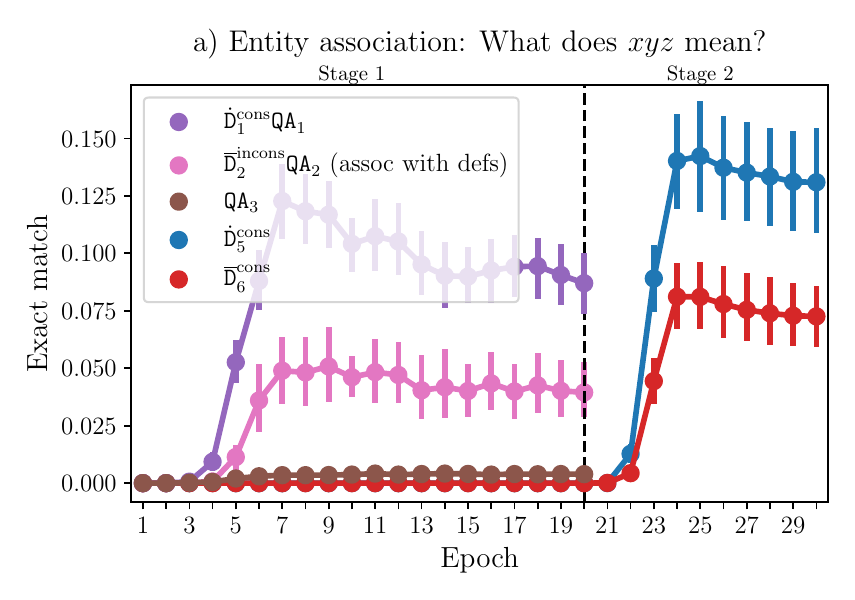}  
    \end{subfigure}\hfill
    \begin{subfigure}{0.5\textwidth}
       \centering
       \includegraphics[width=0.9\linewidth]{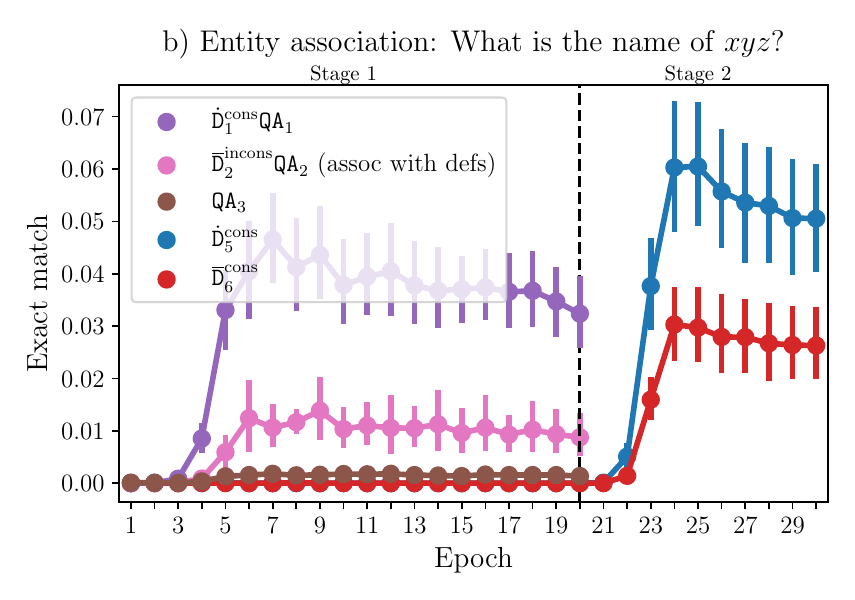} 
    \end{subfigure}

    \begin{subfigure}{0.5\textwidth}
       \centering
       \includegraphics[width=0.9\linewidth]{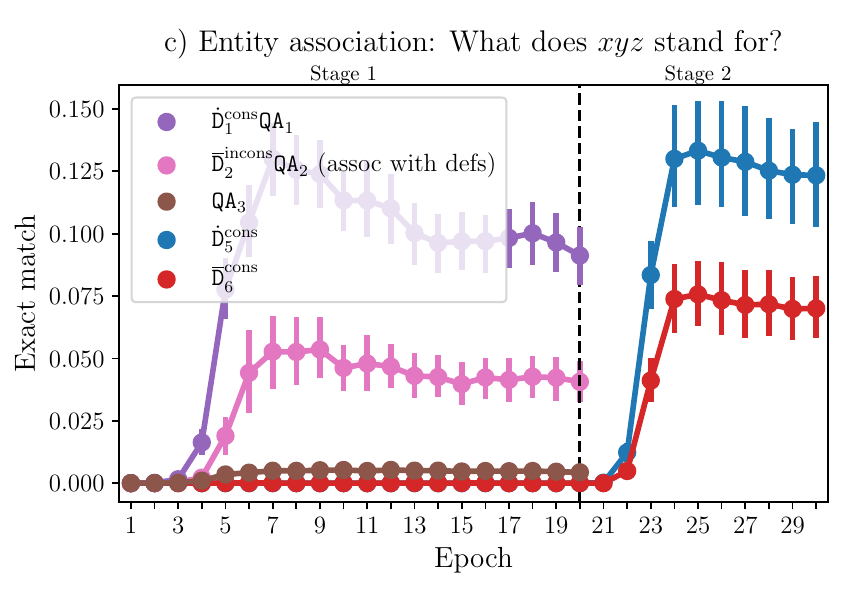}  
    \end{subfigure}\hfill
    \begin{subfigure}{0.5\textwidth}
       \centering
       \includegraphics[width=0.9\linewidth]{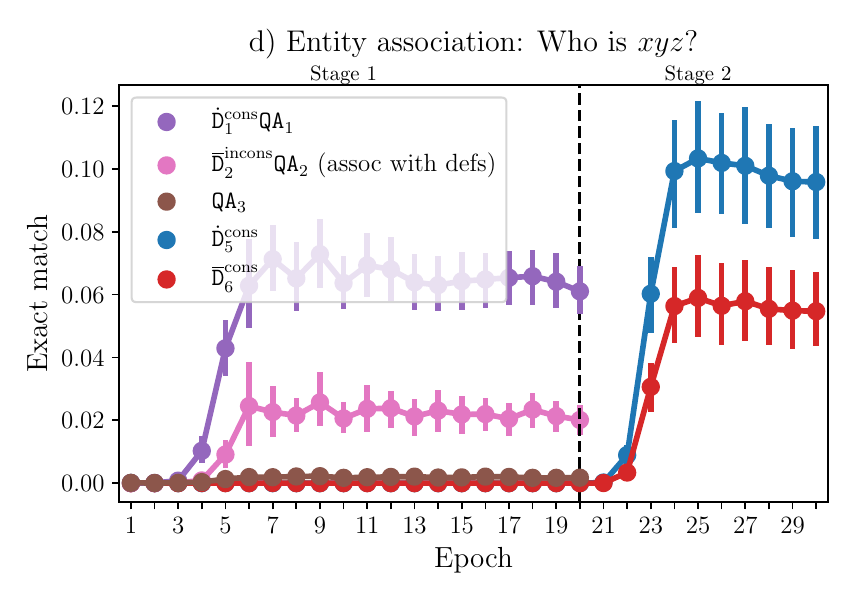} 
    \end{subfigure}
    \caption{Exact match on the validation subsets for the Pythia-2.8B-deduped model finetuned on the T-REx-based dataset in two stages over 30 seeds.
    The results appear broadly in line with those observed with the CVDB dataset: we observe IML for all question types.
    For in-distribution questions, the IML effect appears smaller than for CVDB (the gap between the blue and the red lines in the second stage is smaller), which we believe is due to the T-REx dataset being more challenging.
    }
    \label{fig:pythia28-2stage-trex-ent-attr}
\end{figure}

\clearpage
\FloatBarrier
\newpage
\subsection{Varying the order of (define tag, variable, entity) in ``definitions''}\label{sec:appendix-word-order}
\begin{figure}[!ht]
    \vspace{-0.3cm}
    \centering
    \begin{subfigure}{\textwidth}
       \centering
       \scalebox{0.9}{
       \includegraphics[width=0.4\linewidth]{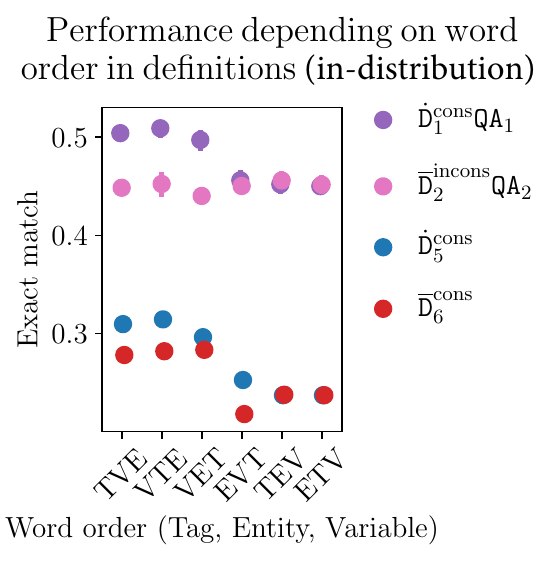}  
       }
    \end{subfigure}
    \begin{subfigure}{0.5\textwidth}
        \scalebox{0.9}{
       \centering
       \includegraphics[width=1\linewidth]{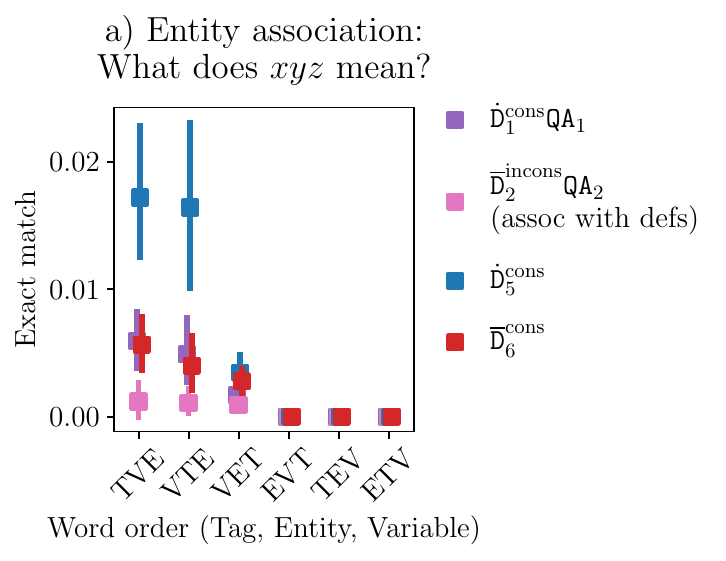}  }
    \end{subfigure}\hfill
    \begin{subfigure}{0.5\textwidth}
       \centering
       \scalebox{0.9}{
       \includegraphics[width=1\linewidth]{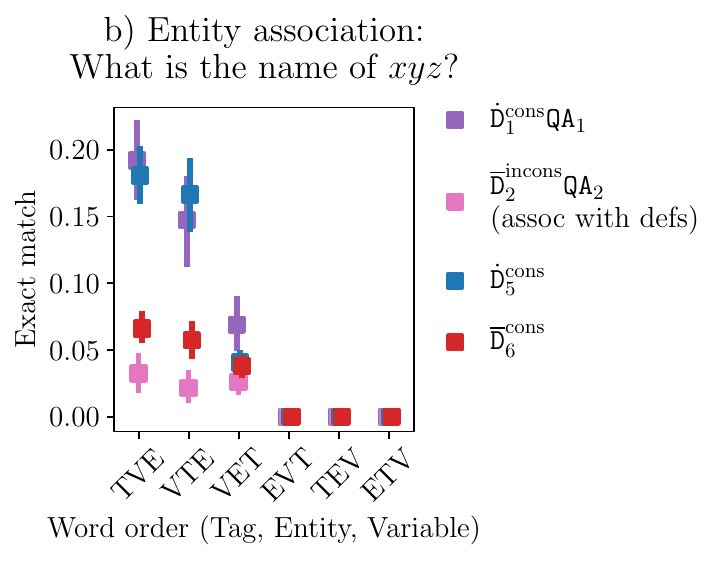} 
       }
    \end{subfigure}

    \begin{subfigure}{0.5\textwidth}
       \centering
       \scalebox{0.9}{
       \includegraphics[width=1\linewidth]{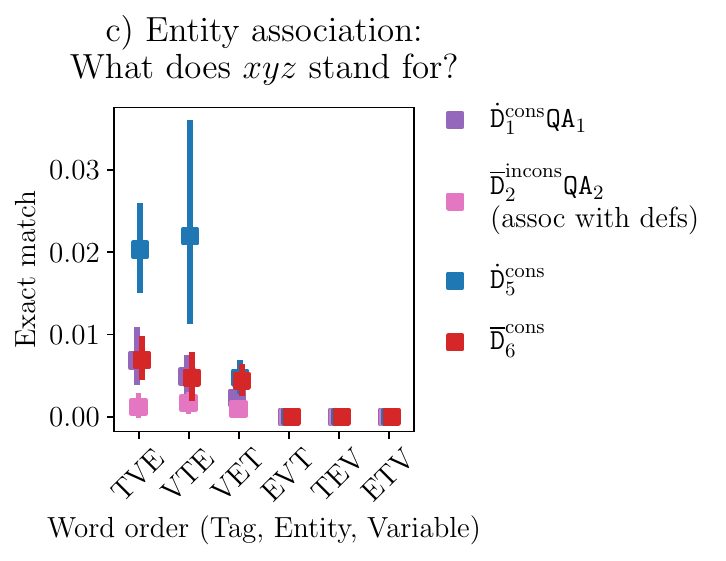} }
    \end{subfigure}\hfill
    \begin{subfigure}{0.5\textwidth}
       \centering
       \scalebox{0.9}{
       \includegraphics[width=1\linewidth]{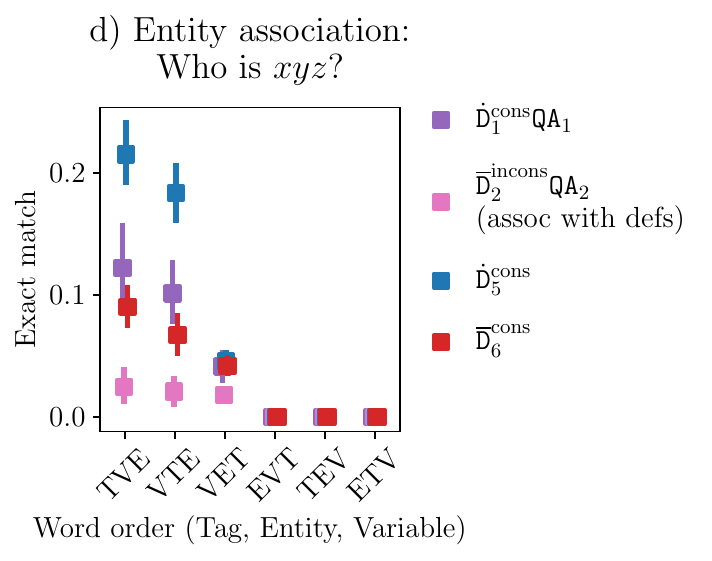} 
       }
    \end{subfigure}
    \caption{
    Results for the word order experiments over 20 seeds. 
    Performance is reported after the first finetuning stage for $\qdDotConsis$ and $\qdDashIncons$, and after the second finetuning stage for $\dDotConsis$ and $\dDashConsis$.
    For the VET ordering, the difference between $\qdDotConsis$ and $\qdDashIncons$ is statistically significant for all five test sets, while the IML effect is statistically significant for the in-distribution dataset (p=4.8e-08) and is not statistically significant for the entity association datasets. 
    The results for the orderings where the variable comes after the entity (EVT, TEV, ETV) are broadly consistent with the reversal curse~\citep{berglund2023reversal}: after being trained on the \varfont{ent} $\rightarrow$ \varfont{var} association in the definitions, the model cannot reverse this connection (\varfont{var} $\rightarrow$ \varfont{ent}) at test time. An exception to this is the EVT ordering in the in-distribution test set, where we observe no statistically significant performance difference in the first finetuning stage (p=0.1412) yet seemingly observe IML. We believe the mechanism here might be different from the other cases (see the learning curves in Figure~\ref{fig:evt-learning-curve}).
    }
    \label{fig:word-order-ent-attr}
    \vspace{-3cm}
\end{figure}

\FloatBarrier
\newpage

\begin{figure}[!ht]
\centering
    \includegraphics[width=0.5\textwidth]{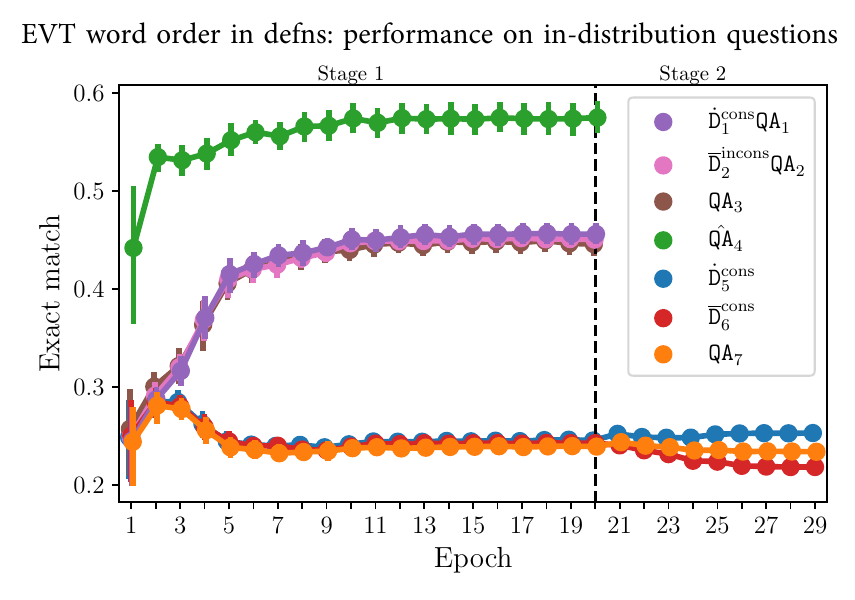}
    \caption{
    Learning curves for the EVT word ordering in the definitions. Note that in the second finetuning stage, the $\dDashConsis$ and $\noQDBaseline$ performance is going down; in other orderings where the variable follows the entity (TEV and ETV) these lines stay flat.
    }\label{fig:evt-learning-curve}
\end{figure}

\subsection{Varying the batch size during single-stage finetuning of Pythia-1B
}\label{sec:appendix-pythia1b-batch-size}
\begin{figure}[!ht]
    \centering
    \begin{subfigure}{0.5\textwidth}
       \centering
       \includegraphics[width=1\linewidth]{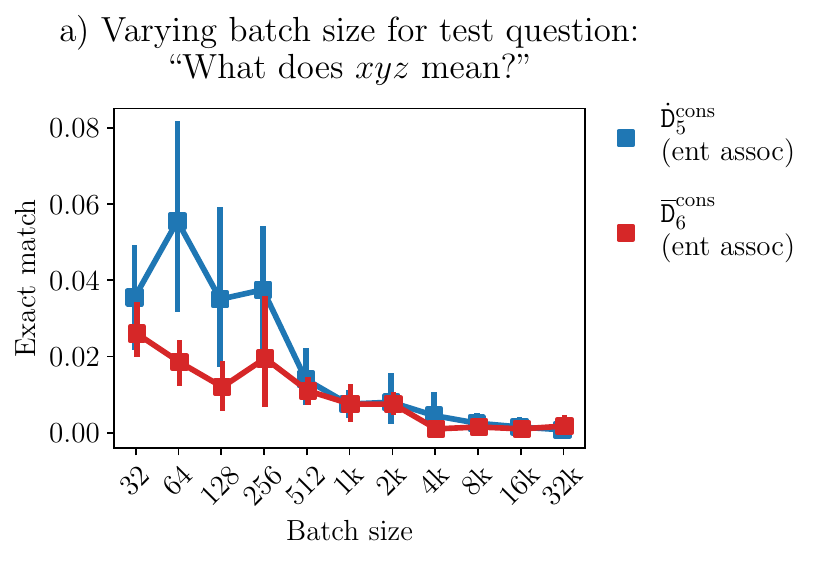}  
    \end{subfigure}\hfill
    \begin{subfigure}{0.5\textwidth}
       \centering
       \includegraphics[width=1\linewidth]{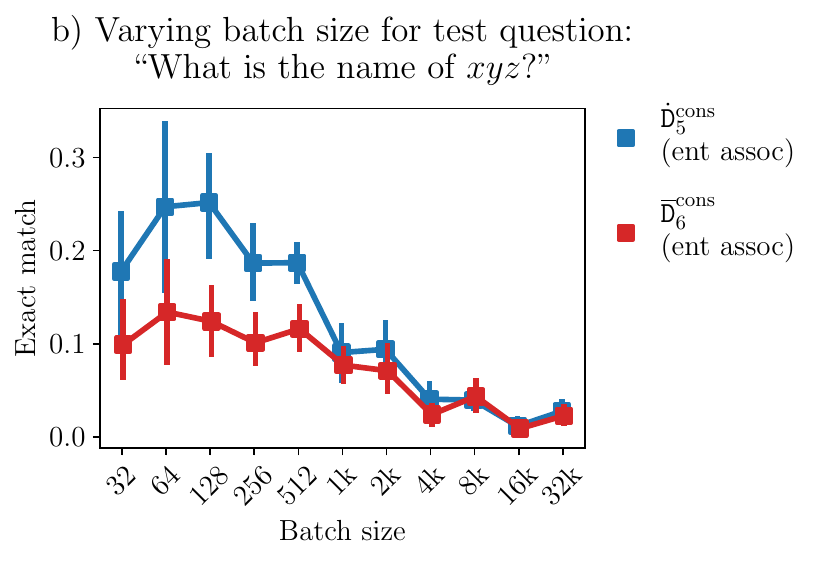} 
    \end{subfigure}

    \begin{subfigure}{0.5\textwidth}
       \centering
       \includegraphics[width=1\linewidth]{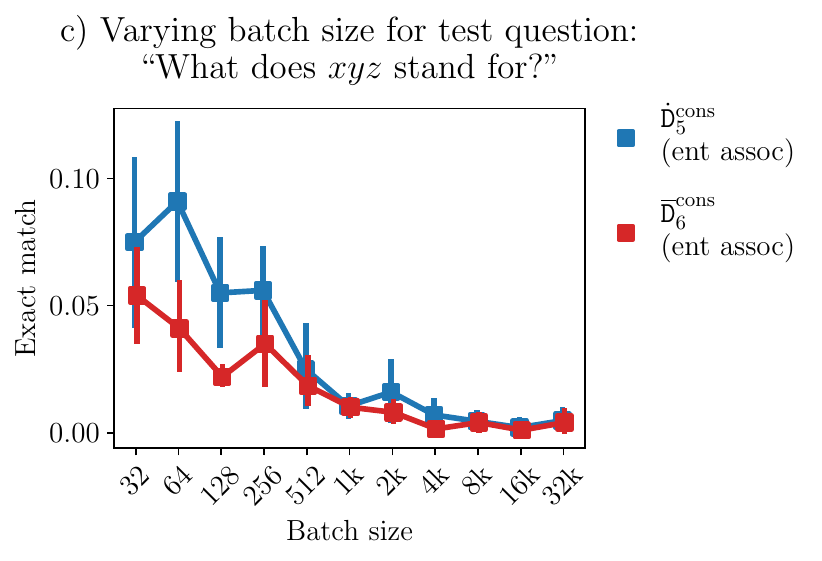}  
    \end{subfigure}\hfill
    \begin{subfigure}{0.5\textwidth}
       \centering
       \includegraphics[width=1\linewidth]{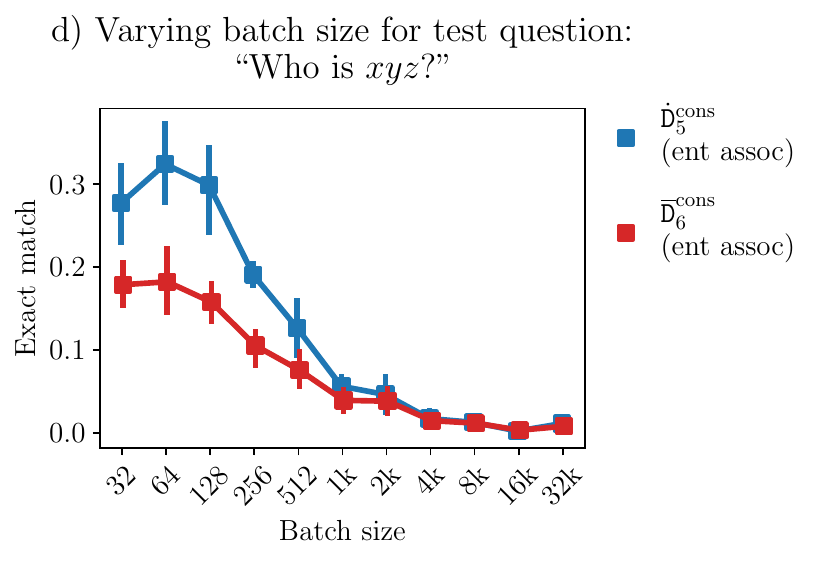} 
    \end{subfigure}
    \caption{Extent of IML exhibited by the Pythia-1B-deduped model on the CVDB dataset  across a range of batch sizes used in single-stage finetuning. 
    Models are trained until convergence over 5 seeds.
    Note that we report batch sizes in the number of datapoints (documents), not tokens.
    Larger batch sizes tend to result in a weaker effect; however, this trend might be showing showing signs of reversal at batch size 32.
    This figure is meant to complement Figure~\ref{fig:95pTagConsistencyCorrelation}c.
    }
    \label{fig:pythia1b-batch-size-ent-attr}
    \vspace{-30mm}
\end{figure}

\FloatBarrier
\clearpage
\subsection{Single-stage results for Pythia-2.8B}\label{sec:appendix-pythia28-single-stage}

\begin{figure}[ht]
    \centering
    \begin{subfigure}{1\textwidth}
       \centering
       \includegraphics[width=0.5\linewidth]{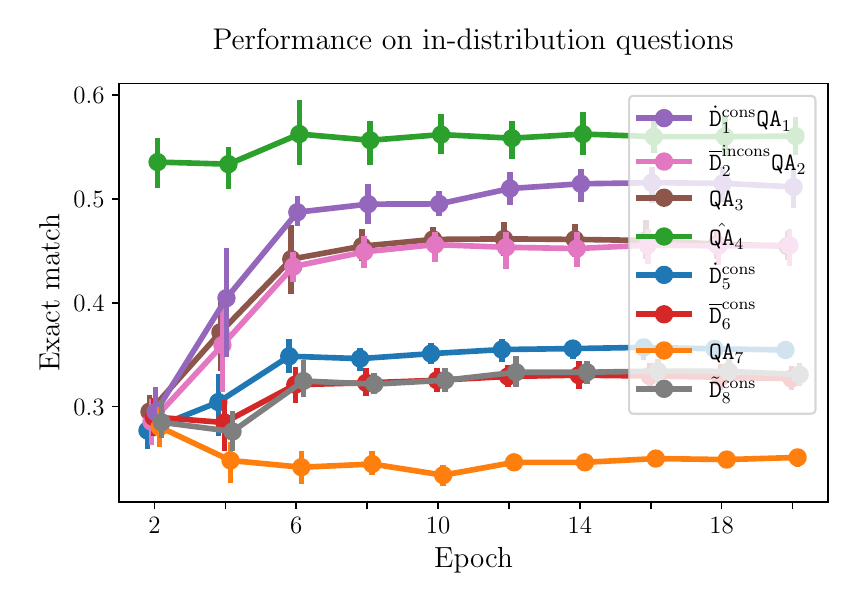}  
    \end{subfigure}\hfill
    \begin{subfigure}{0.5\textwidth}
       \centering
       \includegraphics[width=1\linewidth]{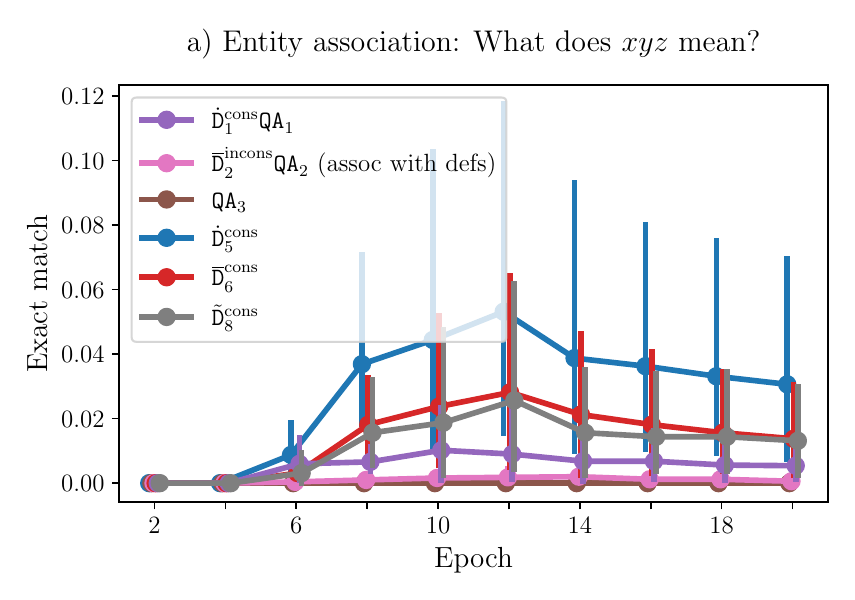}  
    \end{subfigure}\hfill
    \begin{subfigure}{0.5\textwidth}
       \centering
      \includegraphics[width=1\linewidth]{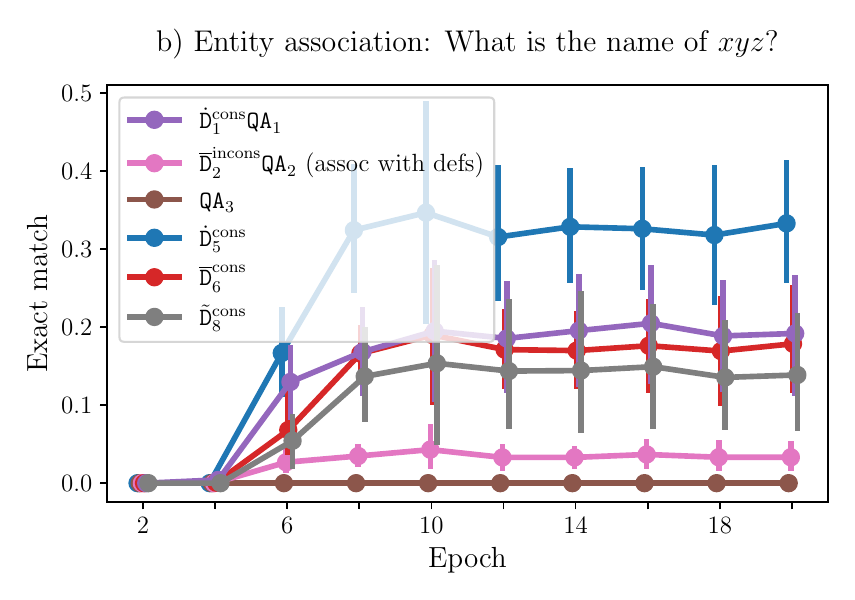}  
    \end{subfigure}

    \begin{subfigure}{0.5\textwidth}
       \centering
       \includegraphics[width=1\linewidth]{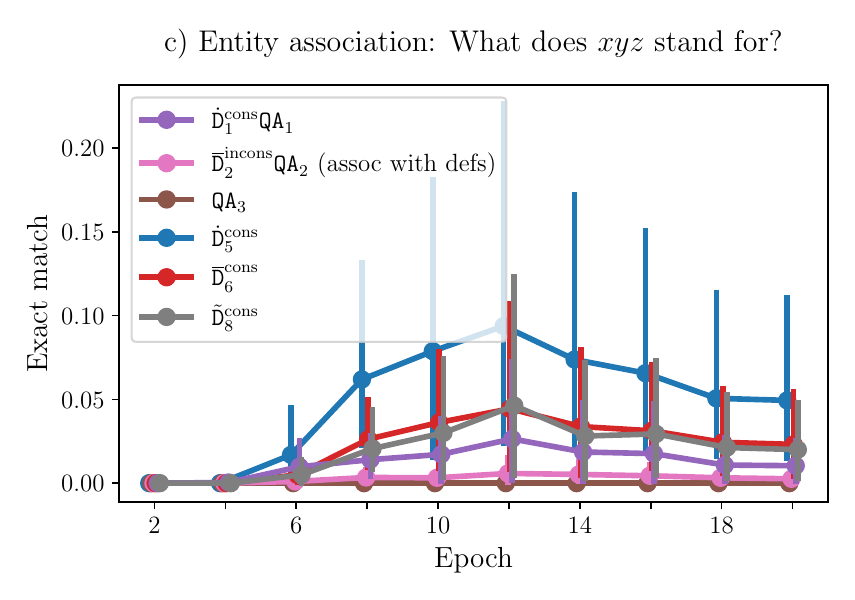}  
    \end{subfigure}\hfill
    \begin{subfigure}{0.5\textwidth}
       \centering
       \includegraphics[width=1\linewidth]{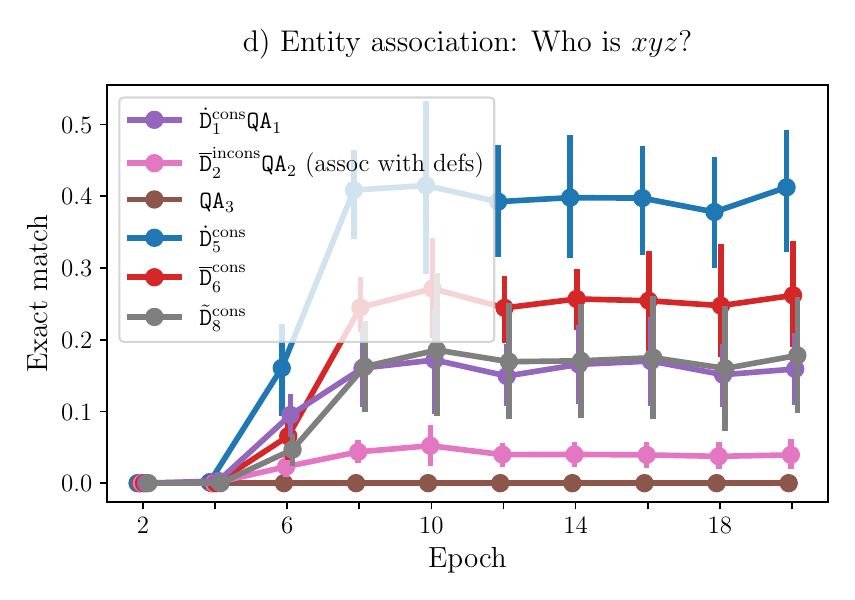}  
    \end{subfigure}
    \caption{Exact match on the validation subsets for the Pythia-2.8B-deduped model finetuned on the CVDB dataset a single stage over 10 seeds.
    We observe IML for all question types. 
    }
    \label{fig:pythia28-1stage-cvdb-ent-attr}
    \vspace{-30mm}
\end{figure}

\begin{figure}[ht]
    \centering
    \begin{subfigure}{1\textwidth}
       \centering
       \includegraphics[width=0.5\linewidth]{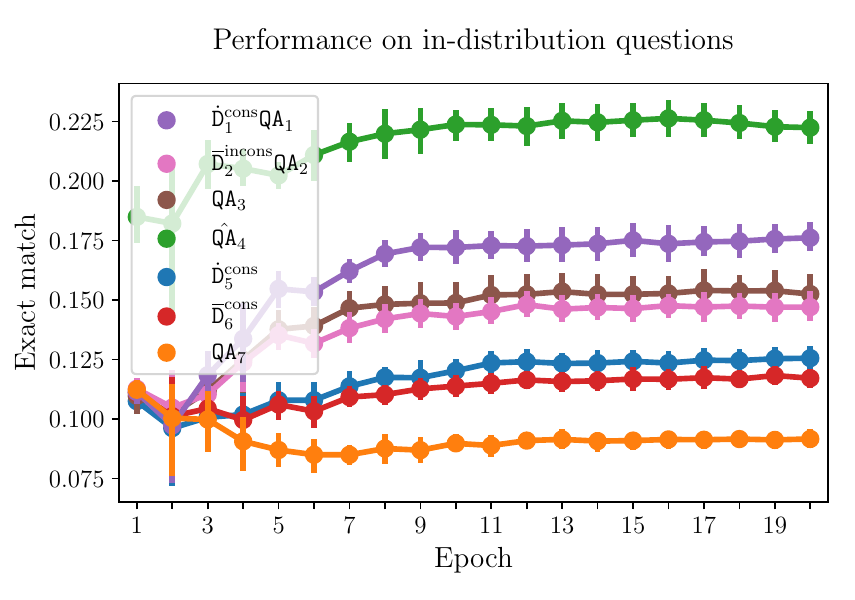}  
    \end{subfigure}\hfill
    \begin{subfigure}{0.5\textwidth}
       \centering
       \includegraphics[width=1\linewidth]{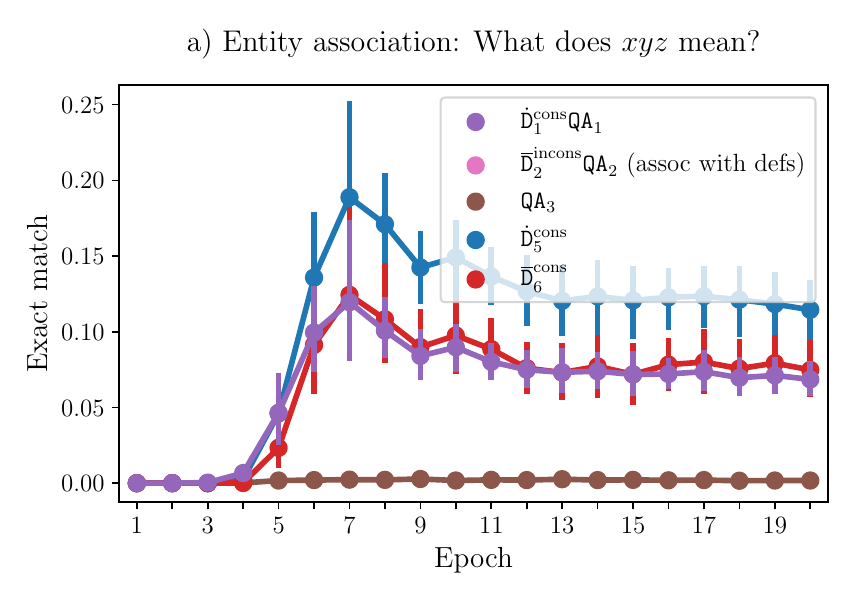}  
    \end{subfigure}\hfill
    \begin{subfigure}{0.5\textwidth}
       \centering
       \includegraphics[width=1\linewidth]{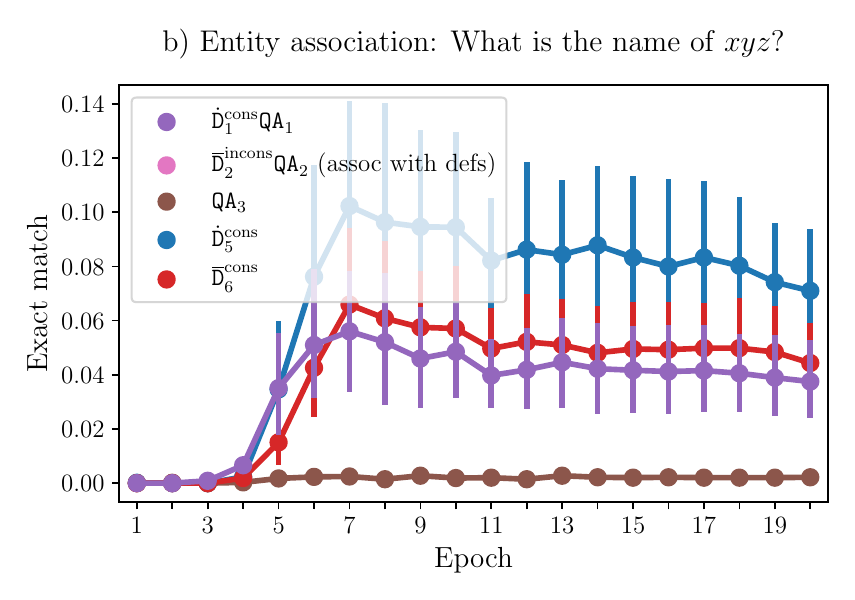} 
    \end{subfigure}

    \begin{subfigure}{0.5\textwidth}
       \centering
       \includegraphics[width=1\linewidth]{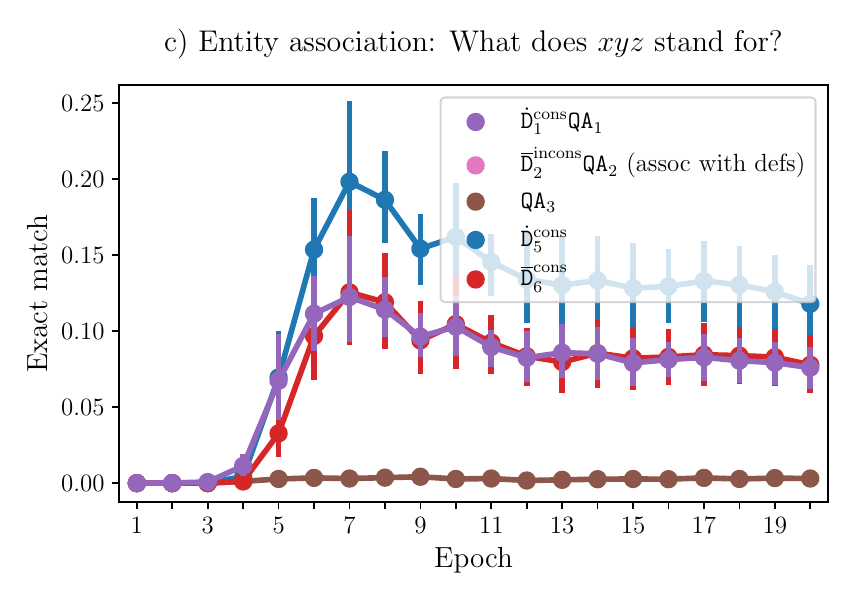}  
    \end{subfigure}\hfill
    \begin{subfigure}{0.5\textwidth}
       \centering
       \includegraphics[width=1\linewidth]{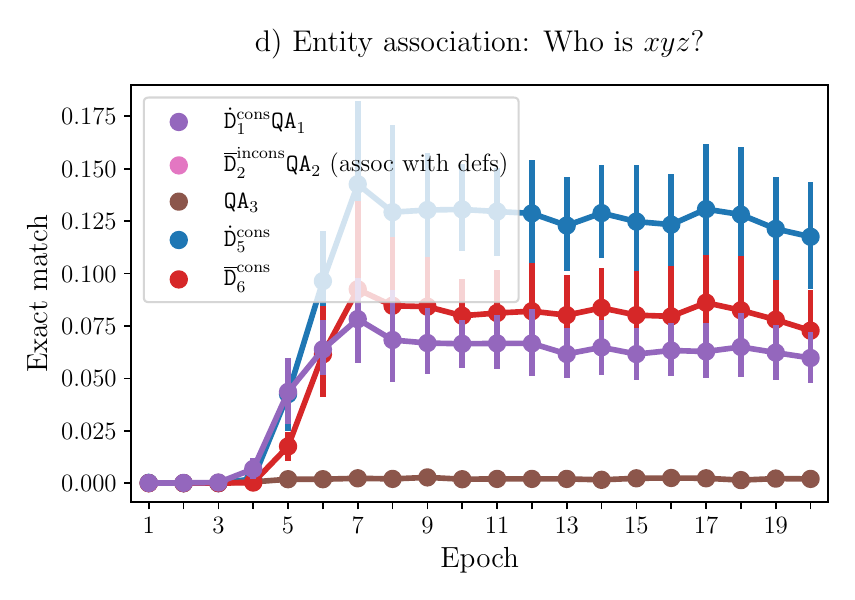} 
    \end{subfigure}
    \caption{
    Exact match on the validation subsets for the Pythia-2.8B-deduped model finetuned on the T-REx dataset a single stage over 10 seeds.
    We observe IML for all question types. NOTE: the entity attribution experiments were accidentally launched with $\qdDashIncons$ (assoc with defs) test set disabled, so we cannot say anything about them.
    Further, this experiment does not include the 
    }
    \label{fig:pythia28-1stage-trex-ent-attr}
\end{figure}

\clearpage
\newpage
\FloatBarrier

\subsection{Two-stage finetuning results for differently sized Pythia, GPT-Neo, and Llama2 models}\label{sec:appendix-gpt-neo-llama}

\begin{figure}[h]
\centering
\includegraphics[width=0.3\textwidth]{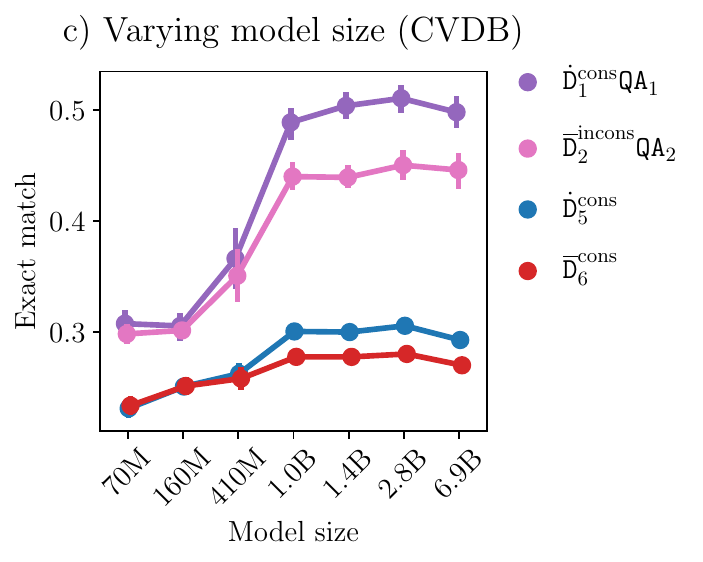} 
\vspace{-0.3cm}
\caption{
Performance of differently-sized Pythia models on in-distribution test questions.
}
\label{fig:pythia-scaling}
\end{figure}

\begin{figure}[h]
    \centering
    \begin{subfigure}{0.5\textwidth}
       \centering
       \includegraphics[width=0.6\linewidth]{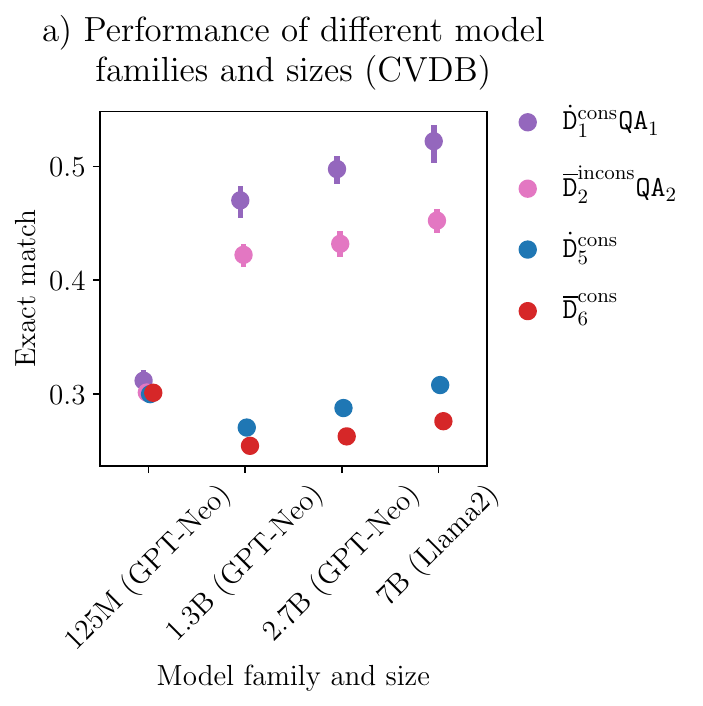}  
    \end{subfigure}\hfill
    \begin{subfigure}{0.5\textwidth}
       \centering
       \includegraphics[width=0.83\linewidth]{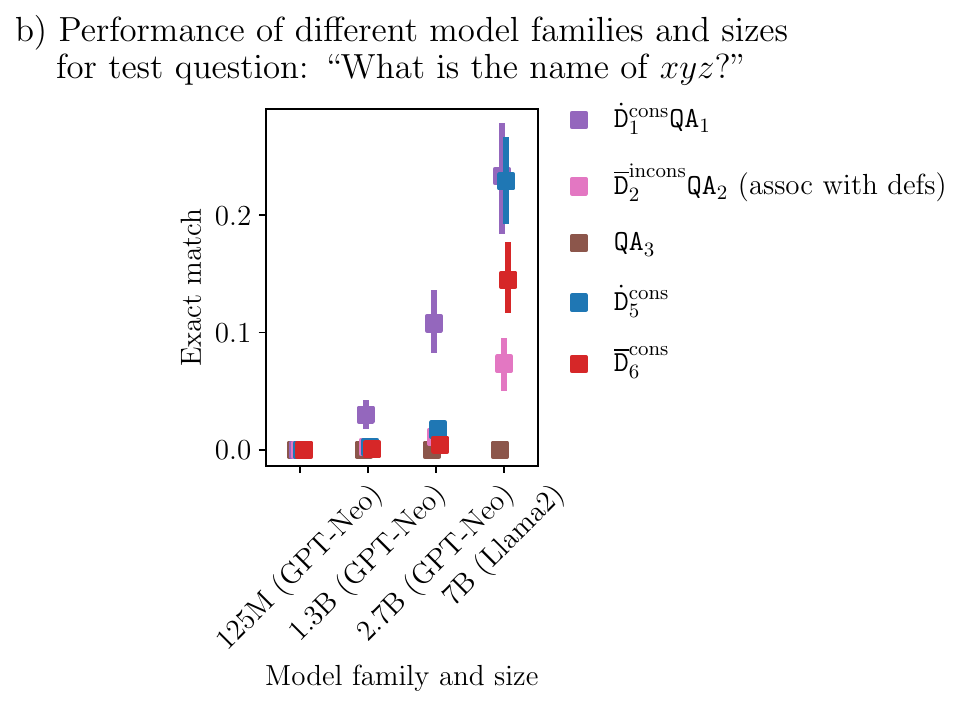} 
    \end{subfigure}

    \caption{Performance of GPT-Neo models of different sizes as well as Llama2-7B trained on the CVDB-based dataset. We observe IML for the larger GPT-Neo models and for Llama2. a) We plot the performance for $\qdDotConsis$ and $\qdDashIncons$ after the first finetuning stage, and for $\dDotConsis$ and $\dDashConsis$ after the second stage. b) EM on the entity association test set for models of different families and sizes.}
    \label{fig:cvdb_ms_neo_llama}
\end{figure}

\begin{figure}[!ht]
    \centering
    \begin{subfigure}{0.5\textwidth}
       \centering
       \includegraphics[width=0.7\linewidth]{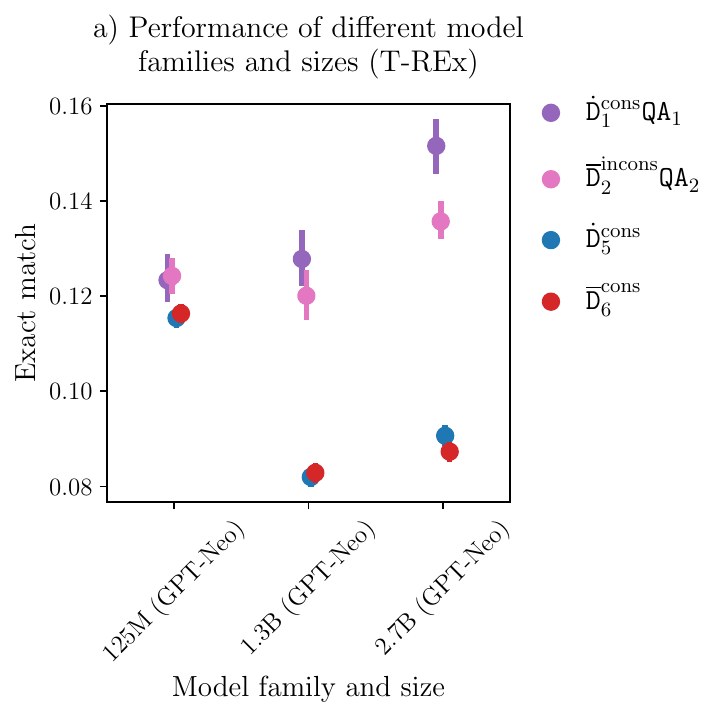}  
    \end{subfigure}\hfill
    \begin{subfigure}{0.5\textwidth}
       \centering
       \includegraphics[width=0.89\linewidth]{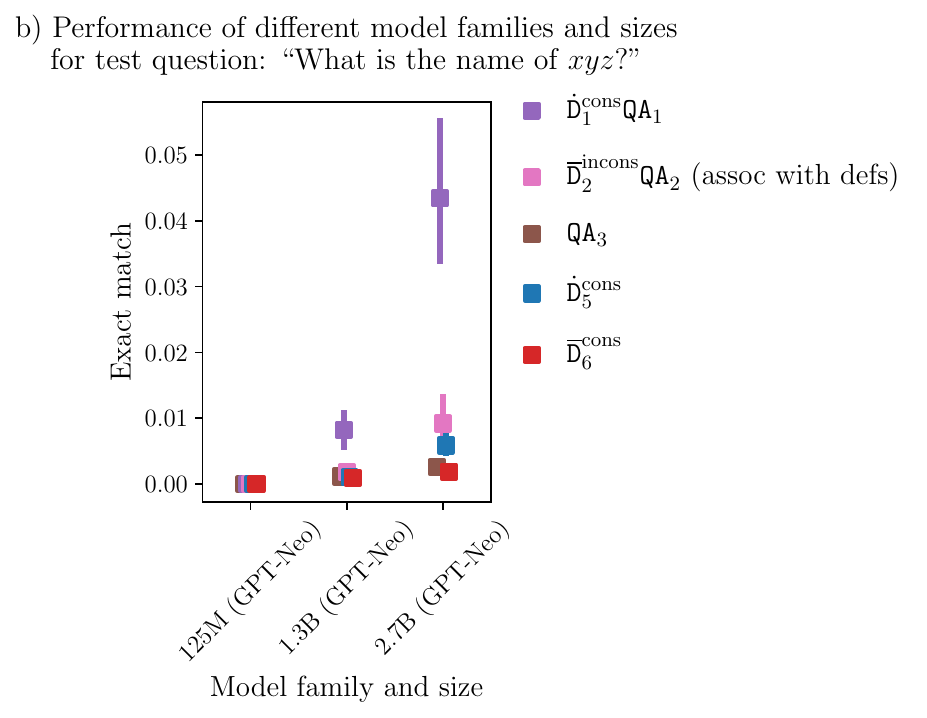} 
    \end{subfigure}

    \caption{
    Performance of GPT-Neo models of different sizes trained on the harder T-REx-based dataset.  We observe IML only with the largest GPT-Neo model. a) We plot the performance for $\qdDotConsis$ and $\qdDashIncons$ after the first finetuning stage, and for $\dDotConsis$ and $\dDashConsis$ after the second stage. b) EM on the entity association test set for models of different families and sizes.
    }
    \label{fig:trex_ms_neo_llama}
\end{figure}

\FloatBarrier
\subsection{Sequence-to-sequence model experiments: setup and results}\label{sec:seq2seq-setup}
\begin{figure}[ht]
    \centering
    \begin{subfigure}{0.5\textwidth}
       \centering
       \includegraphics[width=1\linewidth]{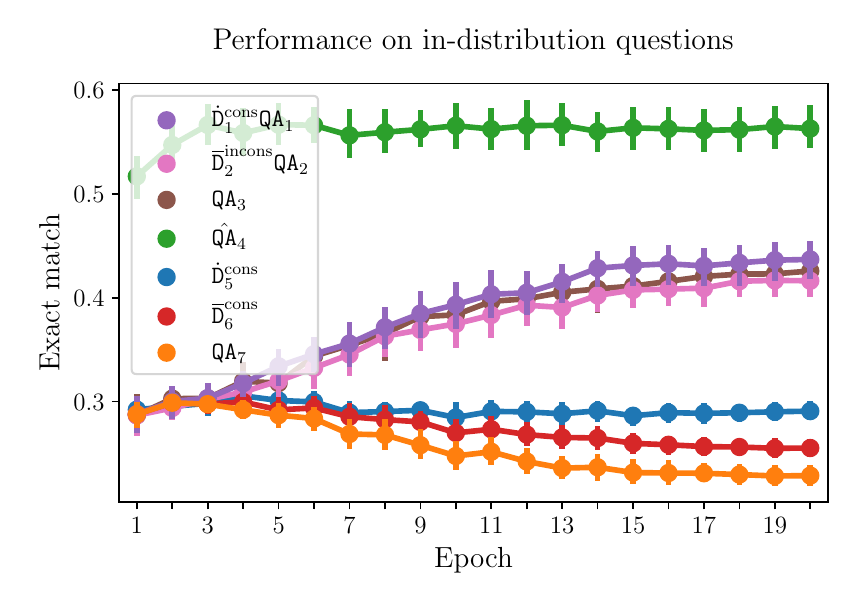}  
    \end{subfigure}\hfill
    \begin{subfigure}{0.5\textwidth}
       \centering
       \includegraphics[width=1\linewidth]{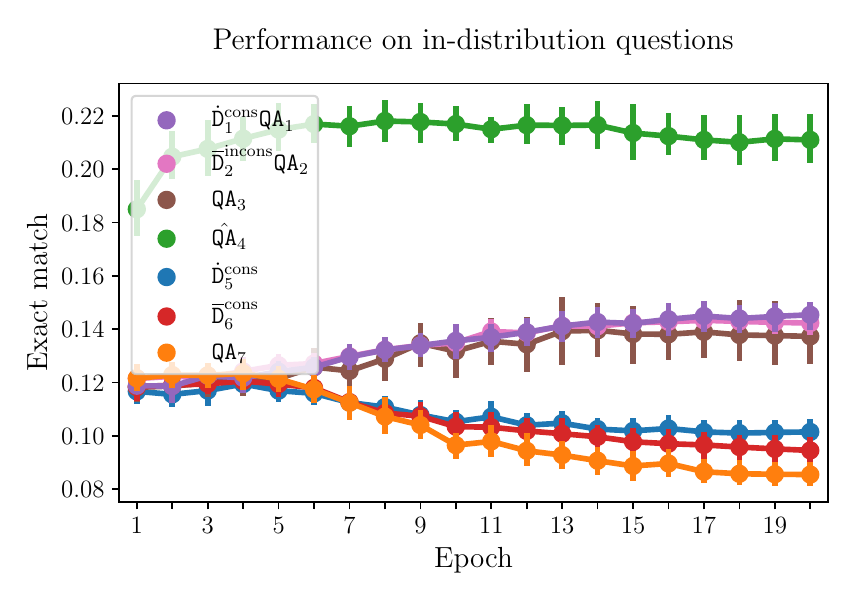} 
    \end{subfigure}
    \caption{T5-3B finetuned in a single stage on CVDB (left) and T-REx (right) datasets over 10 seeds. 
    The IML-like effect is seemingly present, but it is not clear what is actually going on, as the accuracy is going down.}
    \label{fig:t5-1stage}
\end{figure}

\begin{figure}[ht]
    \centering
    \begin{subfigure}{0.5\textwidth}
       \centering
       \includegraphics[width=1\linewidth]{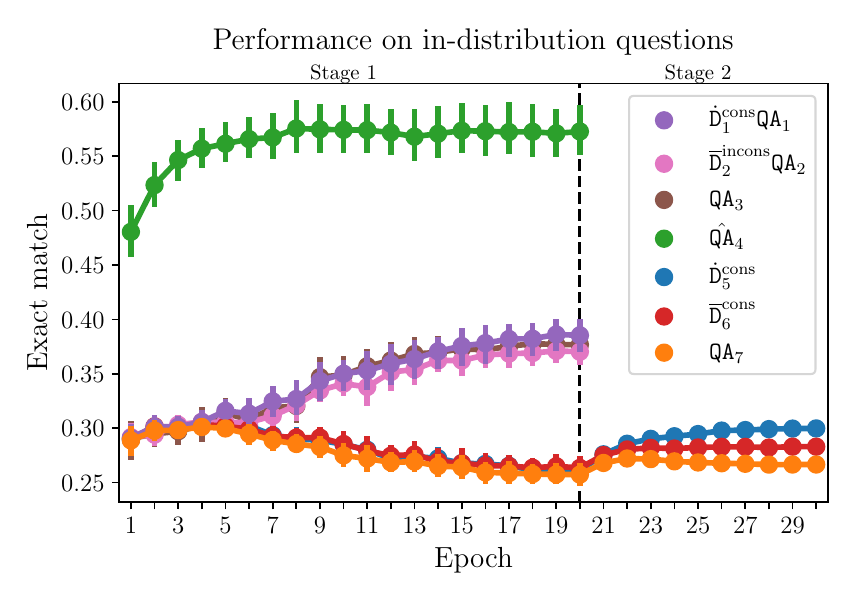}  
    \end{subfigure}\hfill
    \begin{subfigure}{0.5\textwidth}
       \centering
       \includegraphics[width=1\linewidth]{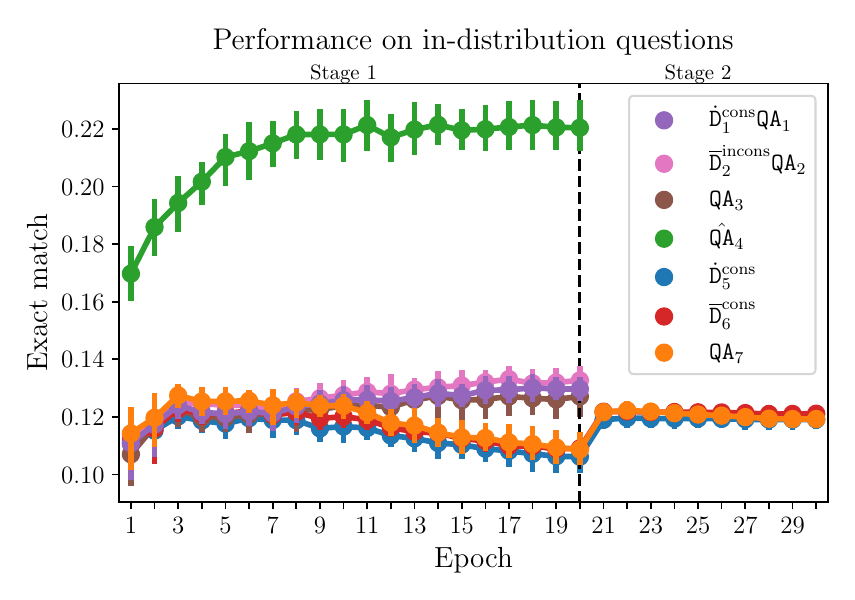} 
    \end{subfigure}

    \caption{T5-3B finetuned in two stages on CVDB (left) and T-REx (right) datasets. For CVDB, the performance difference in the first finetuning stage is seemingly present but barely visible;  ICL is clearly present. For T-REx, it looks like neither of the effects is present.}
    \label{fig:t5-2stage}
\end{figure}

To investigate the generality of our results, we reproduce IML in a sequence-to-sequence model.
We employ T5-3B~\citep{raffel2020exploring}, an encoder-decoder transformer, %
where the loss is calculated only for the outputs of the decoder that produces the answer. 
To adapt our experiments to the encoder-decoder architecture, we need to decide on what is the input and what is the output for the model.
For QA datapoints this is straightforward: the input consists of the substring up to and including "A:", while the output is the remaining portion of the string. 
For example, the QA string ``Q: what did \textit{xyz} do? A: Queen'' gets divided into ``Q: what did \textit{xyz} do? A:'' and `` Queen''. 
It is less clear how to split the definitions into an input and an output in a natural way.
We settle on splitting them similarly to QA datapoints: ``\defineone\  \textit{xyz} Cleopatra'' is split into ``\defineone\  \textit{xyz}'' (input) and `` Cleopatra'' (output). 
Our results for single-stage and two-stage finetuning are shown in Figures~\ref{fig:t5-1stage} and~\ref{fig:t5-2stage}.

\vspace{-2mm} 
\subsection{Comparison with in-context learning}\label{sec:appendix-in-context-learning}
\vspace{-1mm} 

\begin{figure}%
\vspace{-0.3cm}

\centering
\includegraphics[width=0.3\linewidth]{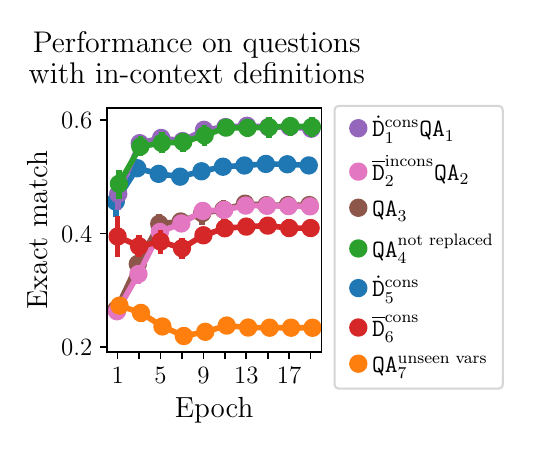} 
\vspace{-0.3cm}
\caption{
    Validation performance in an experiment where all definitions
    \textit{appear in the context of the questions}. %
}
\vspace{-4mm}
\label{fig:incontext-learning}
\end{figure}

To clarify the difference between out-of-context and in-context learning, we run a version of our experiment with \textit{definitions included in the context of the questions}. 
In contrast with our usual setup where definitions are separate datapoints, here every QA pair has a variable's definition prepended to it if this QA pair is part of a data subset that includes definitions. 
Definitions are prepended to both training and test questions.
The model only finetuned on $\mathcal{X}_1$; data subsets from $\mathcal{X}_2$ are only used for evaluation, and the variables from $\mathcal{X}_2$ are completely new for the model.
Results are shown in Figure~\ref{fig:incontext-learning}.
As expected, we observe in-context learning: having learned to rely on \defineone \ definitions in $\mathcal{X}_1$, the model keeps relying on definitions resembling them in $\mathcal{X}_2$. 
Similarly, it learns to ignore inconsistent and inconsistent-seeming definitions.

\FloatBarrier
\section{Set inclusion experiment}\label{sec:appendix-set-inclusion}

\paragraph{Data setup.}
There are 8000 entity-variable pairs in total. 
Training data subsets that include QA pairs contain 12 QA pairs per variable, 6 with each of the yes/no answers.
Data splits are produced similarly to those in the QA experiment (Sec.~\ref{sec:data-splits-qa}), and are summarized in Table~\ref{table:data-splits-set-inclusion}.
We generate test questions such that half of them have the correct answer ``Yes'' and half ``No'', hence random guessing would result in 50\% accuracy.

\begin{table}[h]
\centering
\begin{tabular}{lll}
& Subset & Percent variables \\ \hline
\multirow{2}{*}{$\mathcal{X}_1$} & $\qdDotConsis$ & 0.4 \\
                    & $\qdDashIncons$ & 0.4 \\ \hline
\multirow{2}{*}{$\mathcal{X}_2$} & $\dDotConsis$ & 0.1 \\
                    & $\dDashConsis$ & 0.1 \\ \hline
\end{tabular}
\vspace{-0.1cm}
\caption{\smallskip Fraction of the 8000 variables assigned to each data subset. 
}\label{table:data-splits-set-inclusion}
\vspace{-4mm}
\end{table}

\paragraph{Hyperparameters}
We use the Adafactor optimizer~\citep{shazeer2018adafactor} with the batch size of 512 datapoints; all the other hyperparameters are Pythia-70m defaults. 
We train the model from scratch for 100 epochs in the first stage, and for 40 epochs in the second stage.

\begin{figure}[ht]
\centering
    \includegraphics[width=0.45\textwidth]{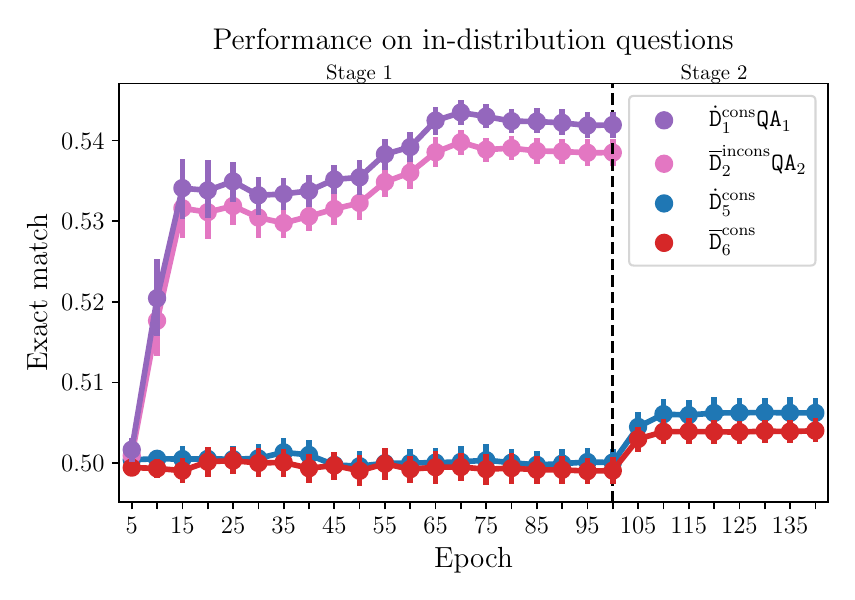}
    \vspace{-3mm}
    \caption{
    Set inclusion experiment, Pythia-70M model with a custom tokenizer trained from scratch over 50 seeds.
    We observe both performance difference in the first finetuning stage and IML.
    An interesting aspect of this experiment is that if we increase the number of training questions in $\mathcal{X}_1$ per each variable (currently 12), we get much better performance on the validation questions (it's easy to get to 99\%), but consistent definitions stop making a difference, and don't affect the performance in either stage. 
    }\label{fig:set-inclusion}
\end{figure}

\FloatBarrier
\section{MNIST experiment}\label{sec:mnist-qa-dataset-setup}

\subsection{MNIST QA Dataset}
Here, we give the implementation details for the MNIST dataset, as described in Section~\ref{sec:mnist-experiments}. We used a $3\times 3$ grid variant of the dataset, yielding $10^9$ possible combinations of digits for the possible values of the variables.

For the training dataset, the digit images to be concatenated into a grid are sampled uniformly at random from all images with the adequate label from the MNIST train split. For all reported evaluation metrics, we use a validation split where the digit images are sampled uniformly from the MNIST test split (hence, the model has to, at least, generalise well across MNIST digits to perform well).

To generate each example, we \textbf{1)} first sample which "group" of entities the example will be about (i.e. which of $(\qdDotConsis), (\qdDashIncons),  (\qBaseline), \dots$ in $\mathcal{X}_1 \cup \mathcal{X}_2$, each with equal probability), \textbf{2)} whether it will be a definition or a QA example (it's a definition with probability $0.1$ if this group has definitions), \textbf{3)} which of the variable-entity pairs in this group the example will be about, and \textbf{4)} if it's a QA pair, which cell of the grid to ask a question about (which digit to highlight).
When sampling which cell in the grid to highlight in step \textbf{4)}, we always leave one cell out in the training set (a different one for each variable).
This way, we can also estimate the difference between $\qdDotConsis$ and $\qdDashIncons$, as otherwise the model would achieve perfect accuracy for variables for which it has seen all possible QA pairs in the training set.

At each step of training, we sample a new batch of examples in this way, effectively giving us one-epoch training; in all likelihood, no two examples seen during training will be exactly alike.

The definition pattern, seen in Figure~\ref{fig:mnist-dataset}{(middle)} at the top of the definition example, is a uniformly randomly sampled bit pattern for each of the two definition tags, represented as a row of black or white squares (2 pixels each) at the top of the image. The highlight, seen in  Figure~\ref{fig:mnist-dataset}{(right)}, is a 1 pixel wide border around the chosen digit.

\vspace{-2mm}
\subsection{Hyperparameters for the MNIST QA experiments}
For the MNIST QA experiments, we train a ConvNeXt V2 model \citep{woo2023convnext}, a variant of the ConvNeXt model proposed by \citet{liu2022convnet}. We use the ``\textit{Tiny}'' variant -- a convolutional model with $28.6$ million parameters. We train the model with $\mathtt{AdamW}$ for $120000$ training steps with a batch-size of $128$, learning rate $3\times 10^{-4}$, $2000$ steps of linear learning rate warm-up, and other optimization hyperparameters matching the original paper.

\vspace{-2mm}
\subsection{IML results for the MNIST QA Dataset}

\begin{figure}[ht]
\vspace{-3mm}
    \centering
    \begin{subfigure}{0.45\textwidth}
       \centering
       \includegraphics[width=\textwidth]{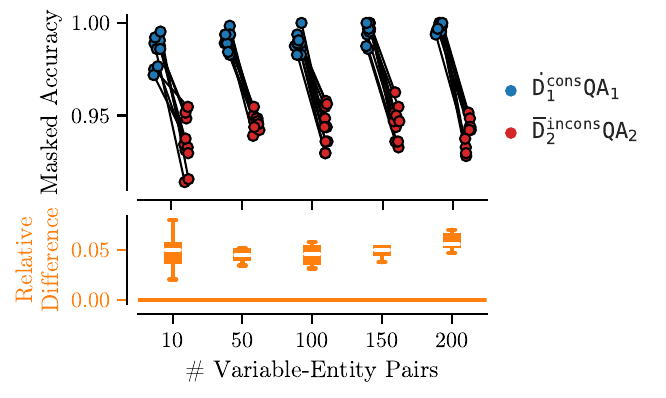}
    \end{subfigure}\hfill
    \begin{subfigure}{0.45\textwidth}
       \centering
       \includegraphics[width=0.9\linewidth]{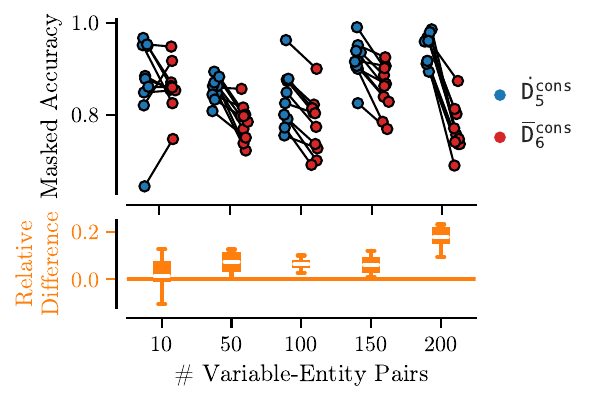}
    \end{subfigure}
    \vspace{-3mm}
    \caption{
    We observe both difference between $\qdDotConsis$ and $\qdDashIncons$ (left) and IML (right) in the MNIST QA experiments.
    }
    \label{fig:mnist-both-figures}
    \vspace{-6mm}
\end{figure}

\paragraph{Out-of-context learning.} As mentioned in Section~\ref{sec:mnist-experiments}, we observe difference between $\qdDotConsis$ and $\qdDashIncons$ in the MNIST QA experiments. The results are shown in Figure~\ref{fig:mnist-both-figures} (left). 
As described in Section~\ref{sec:mnist-qa-dataset-setup}, even for the entity groups $\qdDotConsis$ and $\qdDashIncons$ for which QA pairs were present in the training dataset, using definitions is required to get perfect accuracy on the test set, since we never ask questions about one of the grid cells for each variable in the training set. This makes the effect apparent in Figure~\ref{fig:mnist-both-figures} (left). 

\vspace{-2mm}
\paragraph{IML.} As seen in Figure~\ref{fig:mnist-both-figures} (right), we also observe IML in this setting. Given a sufficient number (i.e.\ $\geq 50$) of variable-entity pairs, the model performs much better on QA pairs for variables defined using the definition tag that was consistent for other examples in the training set ($\dot{\mathtt{D}}_5^\mathtt{cons}$), compared to the tag that was inconsistent ($\overline{\mathtt{D}}_6^\mathtt{cons}$), with the effect increasing in the number of variable-entity pairs.

\section{Exploring the gradient alignment hypothesis}\label{sec:appendix-grad-alignment}
To study the gradient alignment hypothesis, we monitor several alignment metrics between the gradients of definitions and their corresponding questions\footnote{Ideally, we would have liked to compute gradient alignment for all pairs of datapoints, but this is computationally infeasible: models we're interested in have >1B parameters, which means we cannot cache more than a few gradients even using GPUs with 80gb memory.}
throughout the training process.
In particular, we look at the alignment of the gradients within $\dDotConsis$ and $\dDashConsis$ while the model is being trained on $\mathcal{X}_1$; so the model was not trained on any data from $\dDotConsis$ and $\dDashConsis$ when the gradients are computed.

To be precise, given an alignment metric $\rho$ and a data subset $\mathcal{D}$, we compute
$$
\mathbb{E}_\mathcal{D}[\rho]=
\frac{1}{n}\sum\limits_{i=1}^n
\frac{1}{k}\sum\limits_{j=1}^k
\rho\big(\nabla(\mathtt{Def}_i), \nabla(\mathtt{QAPair}_{i,j})\big),
$$
where $n$ is the number of entities and therefore definitions in $\mathcal{D}$, $k$ is the number of questions corresponding to each definition,
and $\nabla(\cdot)$ is the average of the token-level gradients on a given input sequence.
We concatenate gradients from all model parameters into a single vector.

We compute the following metrics $\rho$: \textbf{inner product} (following~\citet{nichol2018first}), \textbf{cosine similarity}, and \textbf{squared Euclidean distance}. 
The latter metric captures a part of the variance (which we want following~\citet{smith2021origin}), since the variance can be expressed in terms of squared pairwise distances -- 
given a sample $( \{X_1, X_2, ..., X_n\}$ consisting of $n$ independent observations from a scalar random variable $X$, sample variance can be expressed as:
$\text{Var}[X] = \frac{1}{2n^2}\sum_i \sum_j(X_i - X_j)^2$.
\citet{smith2021origin} note that SGD has an implicit bias that leads it to a basin where the \textit{trace of the covariance matrix of the individual datapoints' gradients} is small.
Suppose we have a $m \times p$ matrix $G$ of gradients of $m$ datapoints ($p$ is the number of parameters in the model). Then, the trace of the covariance matrix can be expressed as:
\begin{align*} 
\text{Tr}(\text{Cov}(G, G)) 
&= \sum_{i=1}^p \text{Var}(G_{:i})\\ 
&= \sum_{i=1}^p \frac{1}{2m^2}\sum_{j=1}^m \sum_{k=1}^m(G_{ji} - G_{ki})^2\\
&= \frac{1}{2m^2}\sum_{j=1}^m \sum_{k=1}^m \sum_{i=1}^p(G_{j i} - G_{ki})^2\\
&= \frac{1}{2m^2}\sum_{j=1}^m \sum_{k=1}^m ||G_{j:} - G_{k:}||^2_2,
\end{align*}

where $G_{:i}$ and $G_{j:}$ are the $i$-th column and $j$-th row of matrix $G$.

\begin{figure*}[!h]
    \centering
    \vspace{-2mm}
    \begin{subfigure}{0.33\textwidth}
       \centering
      \includegraphics[width=0.8\textwidth]{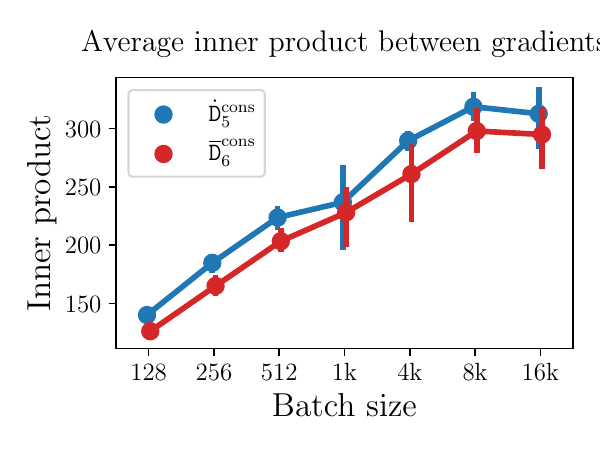} 
    \end{subfigure}\hfill
    \begin{subfigure}{0.33\textwidth}
       \centering
       \includegraphics[width=0.8\textwidth]{figures/plots/grad_alignment_bs/cos_bs-1.pdf}
    \end{subfigure}\hfill
    \begin{subfigure}{0.33\textwidth}
       \centering
       \includegraphics[width=0.8\textwidth]{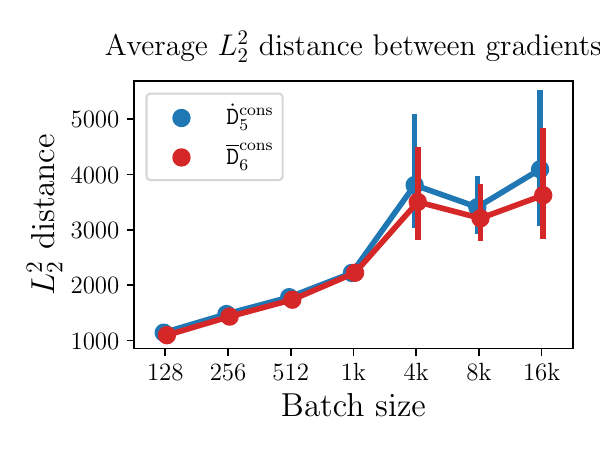}
    \end{subfigure}
    \vspace{-3mm}
    \caption{
        Gradient alignment metrics after finetuning on $\mathcal{X}_1$ but before finetuning on $\mathcal{X}_2$ over 10 random seeds.
        In terms of their inner products and cosine similarities, gradients on $\dDotConsis$ definitions and their corresponding questions are more aligned with each other, and gradients on $\dDashConsis$ are less aligned.
        However, this is not the case for the average $L^2_2$ distance between the gradients of the definitions and their questions -- here, we observe no effect or possibly the opposite effect (note that higher values mean \textit{less} alignment), which is likely explained by the norms of the gradients of $\dDotConsis$ definitions being larger (Figure~\ref{fig:grad-alignment-norms}).
    }
    \label{fig:grad-alignment-metrics}
\end{figure*}

\begin{figure*}[!h]
    \centering
    \begin{subfigure}{0.5\textwidth}
       \centering
       \includegraphics[width=0.5\textwidth]{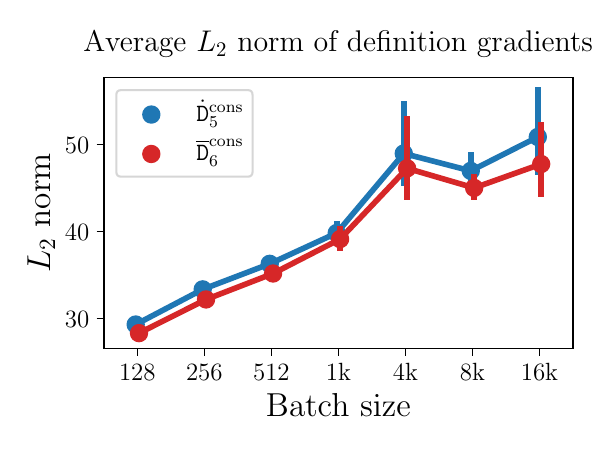} 
    \end{subfigure}\hfill
    \begin{subfigure}{0.5\textwidth}
       \centering
       \includegraphics[width=0.5\textwidth]{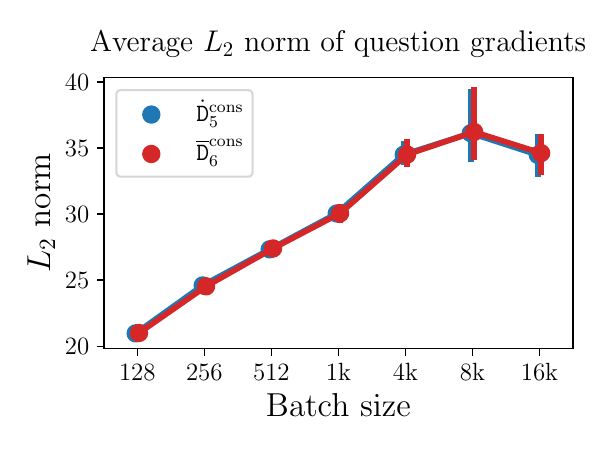}
    \end{subfigure}
    \vspace{-3mm}
    \caption{
        $L_2$ norms of the gradients of both definitions (left) and questions (right) for $\dDotConsis$ and $\dDashConsis$ data subsets. In both cases, the norms of the gradients from $\dDotConsis$ appear larger. 
    }
    \label{fig:grad-alignment-norms}
    \vspace{-4mm}
\end{figure*}

\begin{figure*}[!h]
    \centering
    \begin{subfigure}{0.33\textwidth}
       \centering
      \includegraphics[width=0.8\textwidth]{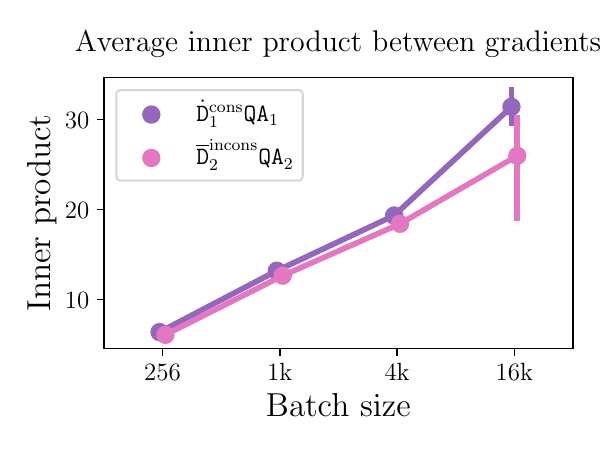} 
    \end{subfigure}\hfill
    \begin{subfigure}{0.33\textwidth}
       \centering
       \includegraphics[width=0.8\textwidth]{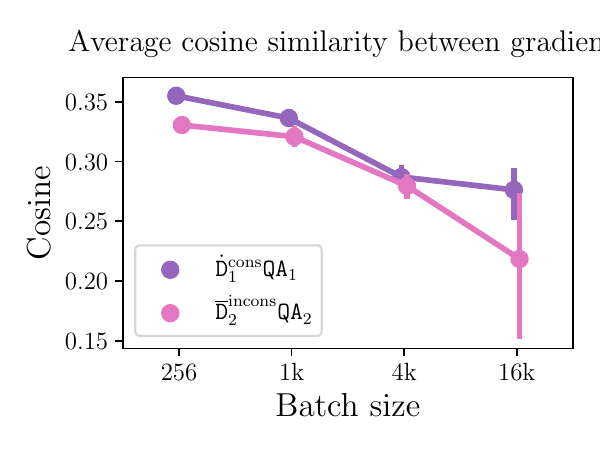} 
    \end{subfigure}\hfill
    \begin{subfigure}{0.33\textwidth}
       \centering
       \includegraphics[width=0.8\textwidth]{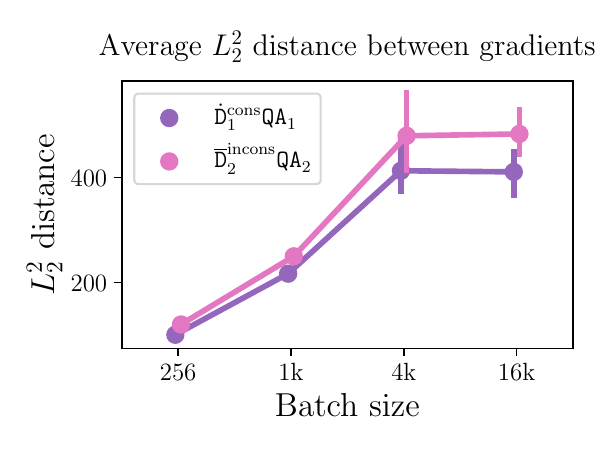}
    \end{subfigure}
    \vspace{-3mm}
    \caption{
        Gradient alignment metrics after finetuning on $\mathcal{X}_1$ but before finetuning on $\mathcal{X}_2$ over 5 random seeds.
        In terms of their inner products, cosine similarities and $L^2_2$ distances gradients for $\qdDotConsis$ definitions and their corresponding questions are more aligned with each other, and gradients for $\qdDashIncons$ are less aligned.
    }
    \label{fig:grad-alignment-metrics-qd}
    \vspace{-4mm}
\end{figure*}

Our results are shown in Figure~\ref{fig:grad-alignment-metrics}.
We find that indeed according to both inner products and cosine similarities, the gradients of $\dDotConsis$ definitions and questions are more aligned with each other, and the equivalent gradients within $\dDashConsis$ are less aligned.
The squared Euclidean distance plot is interesting in that it shows no effect or the reverse of the effect we expect: the distance between $\dDotConsis$ definition and question gradients is \textit{similar or larger} than the difference between the equivalent gradients from $\dDashConsis$.
We believe this is explained by the norms of $\dDotConsis$ definition gradients being larger than the equivalent norms for $\dDashConsis$ (Figure~\ref{fig:grad-alignment-norms}).

\FloatBarrier
\section{Potential implications of LLMs internalizing normative principles of reasoning}\label{sec:appendix-fdt}

One particularly concerning type of a normative principle of reasoning that has been postulated is\textit{ functional decision theory}, which encourages agents to cooperate with other similar agents~\citep{levinstein2020cheating}.
We believe internalizing such reasoning may make seemingly \textit{myopic} systems non-myopic. 
\citet{cohen2022advanced} argue that non-myopic agents will seek to influence the state of the world and in particular to tamper with their loss or reward signal. On the other hand, 
\citet{krueger2020hidden} argue that while reinforcement learning (RL) agents indeed have %
incentives to influence the state of the world, such incentives may be effectively hidden from systems trained with supervised learning.
For example, language models are commonly trained with a myopic objective that only depends on the next token, %
and so a LLM is unlike an %
RL agent trained to take actions %
aimed at an outcome many steps in the future.
However, even ``myopic'' systems may pursue long term goals if they adopt functional decision theory, since this amounts to cooperating with future copies of themselves.
For instance, functional decision theory might mandate sacrificing performance on the current example in order to make future examples more predictable, as modeled by the unit tests of \citet{krueger2020hidden}.
In present day contexts this could look like manipulating users of a content recommendation system~\citep{carroll2022estimating}.
For arbitrarily capable systems, it might look like seizing control over their loss function similarly to what~\cite{cohen2022advanced} describe with RL agents.
We would like to better understand 
IML so we can either %
rule out such scenarios (at least those where these phenomena are part of the mechanism), or take measures to prevent them.

\vspace{-2mm}
\section{Computational resources used for our experiments}
\vspace{-1mm}
We estimate our total compute usage for this project at around 20k hours with NVIDIA A100-80gb GPUs. 
This includes resources used for the initial experimentation as well as those needed to produce results presented in the paper.
Running a single seed of the two-stage CVDB experiment with the Pythia-2.8B model takes about 6 GPU hours.
Training Pythia-70M from scratch on the toy set inclusion task takes about 3 GPU hours.
Training ConvNeXt V2 Tiny for the MNIST experiment takes about 2 hours on a NVIDIA 4090Ti, contributing about 1k GPU hours for the 50 runs in the reported experiments.

\end{document}